\documentclass[journal]{IEEEtran}
\usepackage{amsmath,amsfonts, amssymb}
\usepackage{algorithmic}
\usepackage{algorithm}
\usepackage{array}
\usepackage[caption=false,font=normalsize,labelfont=sf,textfont=sf]{subfig}
\usepackage{textcomp}
\usepackage{stfloats}
\usepackage{url}
\usepackage{hyperref}
\usepackage{verbatim}
\usepackage{graphicx}
\usepackage{cite}
\usepackage{multirow}
\begin{document}

\title{Decoder Gradient Shields: A Family of Provable and High-Fidelity Methods Against Gradient-Based Box-Free Watermark Removal \thanks{The preliminary version was presented at CVPR'2025 \cite{An2025Decoder}.}}

\author{Haonan An,
Guang Hua,~\IEEEmembership{Senior Member,~IEEE,}
Wei Du,
Hangcheng Cao,
Yihang Tao,\\
Guowen Xu,~\IEEEmembership{Senior Member,~IEEE,}
Susanto Rahardja,~\IEEEmembership{Fellow,~IEEE,}
and Yuguang Fang,~\IEEEmembership{Fellow,~IEEE}
\IEEEcompsocitemizethanks{
    \IEEEcompsocthanksitem Haonan An, Yihang Tao, Hangcheng Cao, and Yuguang Fang are with Hong Kong JC STEM Lab of Smart City and Department of Computer Science, City University of Hong Kong, Hong Kong. \protect
    E-mail: haonanan2-c@my.cityu.edu.hk, yihang.tommy@my.cityu.edu.hk, hangccao@cityu.edu.hk,
    my.fang@cityu.edu.hk.
    \IEEEcompsocthanksitem Guang Hua is with the Infocomm Technology Cluster, Singapore Institute of Technology, Singapore 828608. \protect
    E-mail: ghua@ieee.org.
    \IEEEcompsocthanksitem Susanto Rahardja is with the College of Information Science \& Electronic Engineering, Zhejiang University, Hangzhou 310027. \protect
    E-mail: susantorahardja@ieee.org.
    \IEEEcompsocthanksitem Guowen Xu is with the School of Computer Science and Engineering, University of Electronic Science and Technology of China, Chengdu, China. \protect
    E-mail: guowen.xu@uestc.edu.cn.
    \IEEEcompsocthanksitem Wei Du is with the Department of Electronic and Electrical Engineering, University College London, United Kingdom. \protect
    E-mail: uceewdu@ucl.ac.uk.
    }%
\thanks{The research work described in this paper was conducted in the JC STEM Lab of Smart City funded by The Hong Kong Jockey Club Charities Trust under Contract 2023-0108. This work was also supported in part by the Hong Kong SAR Government under the Global STEM Professorship and Research Talent Hub, by Singapore Ministry of Education (MOE) under the Academic Research Fund (AcRF) Tier 1 Grant R-MA123-R205-0008, and by National Natural Science Foundation of China under Grant 62502075.}
\thanks{(Corresponding Author: Hangcheng Cao and Guang Hua.)}
}




\maketitle

\begin{abstract}
Box-free model watermarking has gained significant attention in deep neural network (DNN) intellectual property protection due to its model-agnostic nature and its ability to flexibly manage high-entropy image outputs from generative models. Typically operating in a black-box manner, it employs an encoder-decoder framework for watermark embedding and extraction. While existing research has focused primarily on the encoders for the robustness to resist various attacks, the decoders have been largely overlooked, leading to attacks against the watermark. In this paper, we identify one such attack against the decoder, where query responses are utilized to obtain backpropagated gradients to train a watermark remover. To address this issue, we propose Decoder Gradient Shields (DGSs), a family of defense mechanisms, including DGS at the output (DGS-O), at the input (DGS-I), and in the layers (DGS-L) of the decoder, with a closed-form solution for DGS-O and provable performance for all DGS. Leveraging the joint design of reorienting and rescaling of the gradients from watermark channel gradient leaking queries, the proposed DGSs effectively prevent the watermark remover from achieving training convergence to the desired low-loss value, while preserving image quality of the decoder output. We demonstrate the effectiveness of our proposed DGSs in diverse application scenarios. Our experimental results on deraining and image generation tasks with the state-of-the-art box-free watermarking show that our DGSs achieve a defense success rate of $100\%$ under all settings. 
\end{abstract}


\section{Introduction}
\label{sec:introduction}
\IEEEPARstart{T}{he} importance of intellectual property of deep neural networks (DNNs), especially for generative models, has garnered the attention of governments worldwide \cite{Gov_China_AI, Gov_US_AI, Gov_EU_AI} as the resource invested, including intellectual effort, time, and funding \cite{11127818_Yihang, hucpguard2025_Yihang, tao2025gcpguardedcollaborativeperception_Yihang}, have reached a national scale \cite{Stargate2025}. Among techniques for safeguarding DNNs intellectual property, watermarking
stands out for its general applicability, verifiability, and robustness, which has been adopted in commercial products \cite{google_watermarking_nature, pan2024markllm}.


Based on how watermarks are embedded and extracted, existing DNN model watermarking can be classified into white-box, black-box, and box-free methods \cite{Li2021A_Survey}. White-box watermarks are directly embedded into the model's parameters or architecture, allowing for precise watermark embedding and extraction \cite{Lv2023A_White_Box, Cui2024Steganographic_White_Box,Tondi2024Robust_White_Box}. However, the embedding and extraction processes require access to a model’s internal information, which is highly privacy-sensitive and may violate security protocols. Moreover, this inherent exposure makes white-box watermarking susceptible to watermark removal or tampering, thereby limiting its practical applicability. In contrast, black-box watermarking, first proposed by Adi et al. \cite{Adi2018Turning_Black_Box}, addresses this limitation by introducing a trigger-response mechanism inspired by backdoor attacks (hence, black-box watermarking is also known as backdoor watermarking). Its verification is carried out by observing the model's specific responses to the trigger inputs.

\begin{figure}[!t]
    \centering
    \includegraphics[width=1\linewidth]{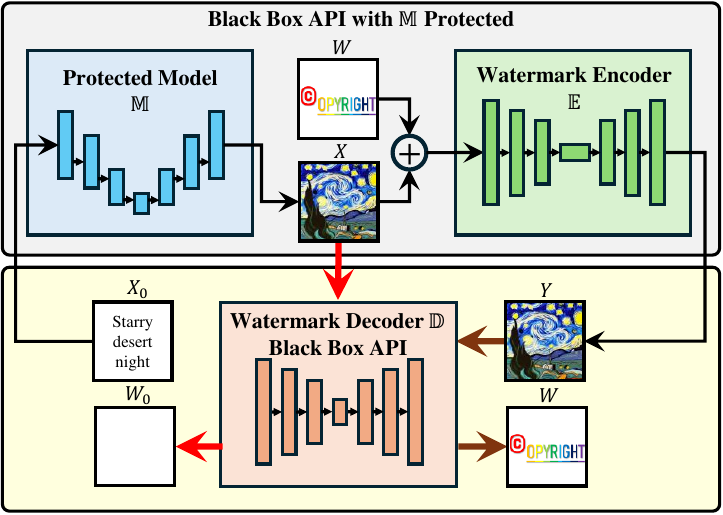}
    \caption{Flowchart of box-free model watermarking for image generative models, which is provided to users via a ``Black-Box API'' (query-based access). The text-to-image generation task is used as an example. Thin black arrows show the black-box querying flow (processing and watermarking), while the thick red arrow represents watermark extraction for a watermark-free image, and the thick brown arrow indicates watermark extraction for a watermarked image.}
    \label{fig:victim_architecture}
    \vspace{-4mm}
\end{figure}

Nevertheless, existing white-box and black-box methods primarily focus on classification models that generate low-entropy outputs, such as probability distributions or the top-1 label, aligning with the nature of watermark embedding into model parameters or behaviors. In contrast, box-free model watermarking enables ownership verification by examining only the model's high-entropy outputs (e.g., images) \cite{Huang2023What_Nonintrusive_GAN}, offering high flexibility and convenience for generative models. The typical workflow of box-free methods is illustrated in \mbox{Fig. \ref{fig:victim_architecture}}, where text-to-image generation task is considered as an example. Notably, the term ``Black-Box API'' in \mbox{Fig. \ref{fig:victim_architecture}} refers to the query-based access level to the system, which is distinct from ``black-box watermarking''. The text-to-image model, denoted by $\mathbb{M}$, takes an input $X_0 \in \mathcal{X}_0$ and generates an output $X \in \mathcal{X}$. Then, the watermark encoder $\mathbb{E}$ embeds a copyright watermark image $W$ into $X$ and generates $Y \in \mathcal{Y}$ visually similar to $X$. To verify the watermark, a dedicated decoder $\mathbb{D}$ is used, which can extract $W$ from the watermarked image set $\mathcal{Y}$ or a null watermark $W_0$ from the non-watermarked set $\mathcal{Y}^\complement$. Depending on the task at hand, the pair $(\mathcal{X}_0, \mathcal{X})$ could represent various transformations, such as (text, image), (noisy image, denoised image) or (original style image, transferred style image). Throughout this procedure, the model $\mathbb{M}$ is safeguarded by only releasing the watermarked output $Y$ rather than the original output $X$. Prior research has shown that even when an attacker uses both $X_0$ and $Y$ to train a surrogate model, $\mathbb{D}$ can still extract the watermark $W$ from the surrogate's output images \cite{Zhang2022Deep_Box_Free}. For convenience, $\mathbb{M}$ is referred to as the protected or victim model, depending on the context.

\begin{figure}[!t]
    \centering
    \includegraphics[width=1\linewidth]{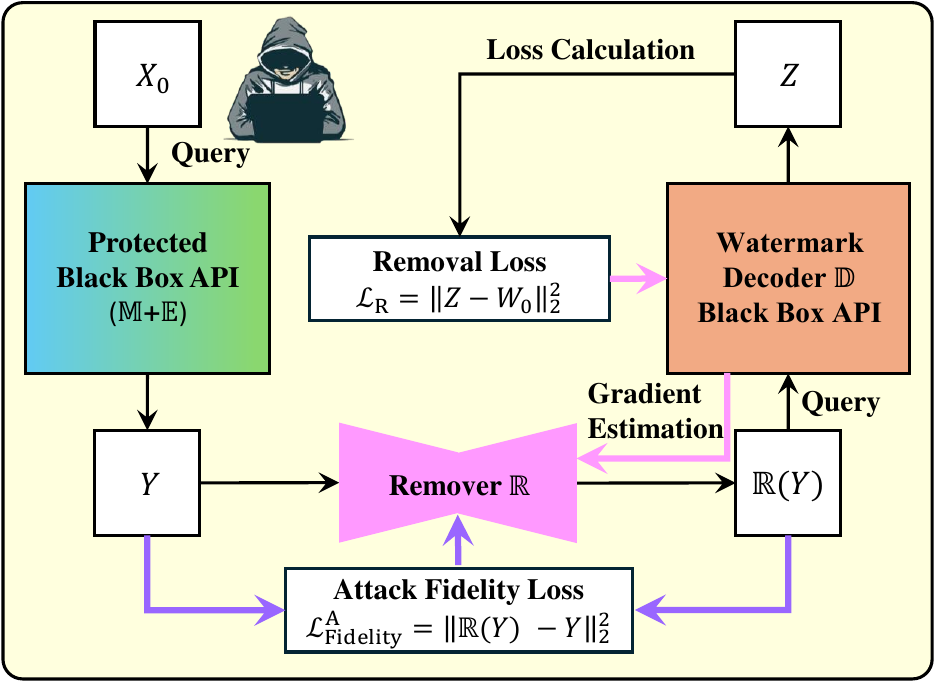}
    \caption{Flowchart of gradient-based removal attack. The gradient backpropagated from $\mathbb{D}$ can be estimated by leveraging black-box adversarial attacks. However, to demonstrate the effectiveness of our proposed methods, we consider an attacker-friendly setting which assumes the attacker can directly obtain the gradient without estimation.}
    \label{fig:gradient_based_attack}
    \vspace{-3mm}
\end{figure}

An attacker can modify $Y$ in an attempt to remove the watermark while maintaining the image quality, prior to training a surrogate model. Potential modifications may include actions such as compression, noise addition or filtering, flipping, cropping, and more. While such vulnerabilities can be partially addressed by incorporating an augmentation layer between $Y$ and $\mathbb{D}$ during the training of both $\mathbb{E}$ and $\mathbb{D}$ \cite{Zhang2024Robust_Box_Free}, the attacker can still perform a more sophisticated removal attack by training a dedicated removal network $\mathbb{R}$, as illustrated in Fig. \ref{fig:gradient_based_attack}. Research on adversarial attacks \cite{ilyas2018black_adversarial, Dong2022Query_adversarial, shi2022query_adversarial} indicates that the gradient of $\mathbb{D}$ can be approximated through black-box queries, which can then be utilized to train $\mathbb{R}$. This gradient-based attack is possible because $\mathbb{D}$, being jointly trained with $\mathbb{E}$, inherently contains the watermarking information, which can be counteracted. Additionally, such an attack requires the inclusion of a fidelity loss alongside the removal loss during the training of $\mathbb{R}$, to preserve image quality.

In this paper, we begin by formulating the gradient-based watermark removal attack discussed above. To comprehensively address this threat, we propose Decoder Gradient Shields (DGSs), a family of defense methods that act as protective layers inserted at the $\mathbb{D}$'s output, input, or intermediate layers, with provable performance. We demonstrate that, even in the setting that favors the attacker, DGSs are capable of disrupting the attack by either reorienting or rescaling the gradients, thereby preventing the remover $\mathbb{R}$ from reaching convergence to the desired low-loss value (with a slight abuse of terminology, hereafter referred to as ``preventing convergence'' for brevity). To be specific, based on the applied location, we categorize DGSs into: \mbox{\textbf{1) DGS at the Output (DGS-O):}} A neat closed-form solution to generate perturbations applied to the decoder's output, with high computational efficiency. \mbox{\textbf{2) DGS at the Input (DGS-I):}} Defensive perturbations applied to the input of the decoder. With the orthogonality-based perturbation selection mechanism, DGS-I significantly enhances the security level compared with DGS-O but incurs additional computational burden. \mbox{\textbf{3) DGS in the Layer (DGS-L):}} This method also employs orthogonal perturbations, but applies them to the intermediate decoder layer, enhancing both efficiency and robustness over DGS-I. However, it requires access to $\mathbb{D}$’s internal information, which could be privacy-sensitive in some cases. Each method contains its own advantages and disadvantages, allowing the defender to choose according to specific application scenarios. Moreover, all three methods are capable of preserving the image quality of $\mathbb{D}$'s output, with only negligible impact on normal user experience. To summarize, the contributions of this paper are listed as follows
\begin{itemize}
    \item We reveal a vulnerability in the decoder of box-free model watermarking: the decoder can be exploited through black-box access, offering an opportunity for the training of a watermark remover.
    \item We propose DGSs, a family of protection mechanisms, including DGS-O, DGS-I, and DGS-L, as countermeasures against gradient-based watermark removal for box-free model watermarking, while preserving image quality.
    \item We present detailed derivations for all DGSs, including the closed-form solution for DGS-O, and orthogonality-based generation processes for DGS-I and DGS-L. 
    \item We thoroughly analyze the respective strengths and weaknesses of DGS-O, DGS-I, and DGS-L across diverse defense scenarios, and conduct extensive experiments for image-to-image and text-to-image tasks to demonstrate their effectiveness.
\end{itemize}

This paper is the extended version from our preliminary results \cite{An2025Decoder} that focused only on DGS-O, and herein it provides the following in-depth and additional contents. 1) We provide more detailed analysis of the choice of the matrix $P$ in DGS-O and summarize two guiding design mechanisms for practice called Gradient Norm Reduction and Randomness Injection. 2) We discuss the underlying vulnerability of DGS-O and present the details of two novel DGSs, i.e., DGS-I and DGS-L, with theoretical analysis. 3) We extend all the experimental data from grayscale images \cite{An2025Decoder} to RGB images. 4) We provide additional experimental results on DGS-I and DGS-L, text-to-image task, watermark sensitivity, and computational cost to further demonstrate the effectiveness and applicability of our proposed methods.


\section{Related Work}
\subsection{Box-Free Model Watermarking}
\label{sec:boxfree}
Box-free model watermarking stands out for its flexibility, embedding watermarks directly into model outputs and employing a dedicated decoder for extraction. This design aligns well with the pressing demand for protecting the intellectual property of generative models and corresponding outputs \cite{Li2021A_Survey}. Yu et al. \cite{yu2021artificial, yu2022responsible} introduced two distinct methods for securing generative adversarial networks (GANs), where watermarks were embedded into the training data and the generator, respectively, with extraction performed on the generated images. In addition, Wu et al. and Lukas et al. \cite{Wu2021Watermarking_Box_Free,Lukas2023PTW_Box_Free} adopted a similar idea from \cite{yu2022responsible} by jointly training both the model and the watermark decoder, but for image-to-image models. Lin et al. and Fei et al. \cite{Lin2024A_Box_Free,Fei2024Wide_Box_Free} proposed more efficient frameworks by fine-tuning the protected models while keeping the pretrained decoder frozen. Moreover, the methods proposed by Zhang et al. \cite{Zhang2022Deep_Box_Free,Zhang2024Robust_Box_Free}, which adopt traditional image watermarking techniques to model watermarking by decoupling the model operation and the watermark embedding, achieve the state-of-the-art performance, which will be considered as victim models in our experiments.

\subsection{Watermark Removal}
For box-free model watermarking, the primary objective for watermark removal is to eliminate the transferable ownership claim, i.e., the watermark, embedded in the victim model for untraceable model extraction, also known as surrogate attack or model stealing. Surrogate attack leverages the queries to the victim model and the corresponding outputs to train a surrogate model to replicate the victim model's functionality for further benefit \cite{Barbalau2020Black, Kariyappa2021Maze, Ma2023DivTheft, Sha2023Cant}. To this end, attackers may manipulate the victim model’s outputs through common image transformations \cite{Zhang2022Deep_Box_Free} or apply specialized operations such as inpainting \cite{Liu2023Erase_Removal_Attack}. However, these approaches are typically inspired by traditional image watermark removal techniques and ignore the inherent coupling between the encoder and decoder in box-free model watermarking. As a result, their effectiveness against such systems remains limited \cite{Zhang2024Robust_Box_Free}.

\subsection{Gradient Attack and Defense}
Previous attacks on the gradients of DNNs have primarily focused on generating adversarial examples, which mislead models into producing unintended outputs. In white-box settings, classic methods such as the fast gradient sign method (FGSM) \cite{goodfellow2014explaining} and projected gradient descent (PGD) \cite{madry2018towards} directly utilize model gradients to craft perturbations. In more practical black-box scenarios, gradient information can be estimated through model queries \cite{ilyas2018black_adversarial}, and the number of required queries can be significantly reduced through improved techniques \cite{Dong2022Query_adversarial}. These advancements form the foundation of our threat model, which assumes that the attacker has access to the gradients of the black-box decoder. In the field of model watermarking, gradient manipulation has primarily been used as a defensive mechanism to prevent extraction of classification models \cite{Mantas2022How, pp_2020_iclr_defense_model_stealing}. However, such defenses are ineffective against hard-label-based extraction attacks \cite{Sanyal2022Towards_Steal_Attack}, and have not been extended to generative models. Moreover, they often have high computational costs due to the complexity of their optimization procedures. Our work here is the first to introduce gradient modification as a protection mechanism tailored specifically for box-free watermarking in generative models, offering a family of effective methods to protect intellectual property of DNNs. 

\begin{table}[!t]
\renewcommand{\arraystretch}{1.0}
\centering
\caption{List of Important notations.}
\label{tab:notation} 
\setlength{\tabcolsep}{4pt}
{\begin{tabular}{c|c}
\hline
\hline
Notation  & Definition \\
\hline
DGSs & Decoder Gradient Shields\\
DGS-O & DGS applied at the decoder's output \\
DGS-I & DGS applied at the decoder's input \\
DGS-L & DGS applied in an intermediate layer of the decoder \\
$\mathbb{M}$ & The protected model \\
$\mathbb{E}$ & The watermark encoder \\
$\mathbb{D}$ & The watermark decoder \\
$\mathbb{R}$ & The attacker's watermark removal network \\
$X_0$ & Initial input to the model $\mathbb{M}$ (e.g., text prompt) \\
$X$ & Original, non-watermarked output from $\mathbb{M}$ \\
$W$ & The true copyright watermark image \\
$Y$ & The watermarked output image produced by $\mathbb{E}$ \\
$W_0$ & The null watermark (representing absence of $W$) \\
$S$ & An arbitrary image query to the decoder $\mathbb{D}$ \\
$Z$ & $Z \triangleq \mathbb{D}[\mathbb{R}(Y)]$ \\
$Z^{\ast}$ & Modified (perturbed) output from DGS-O \\
$P$ & Positive definite matrix used in DGS-O \\
$\mathbb{D}^\ast(S)$ & API response from the DGS-O-protected decoder \\
$\Tilde{Z}$ & Attacker's estimation of $Z$ \\
$\tilde{S}$ & Perturbed $S$ in DGS-I\\
$\eta(S)$ & Adversarial perturbation added in DGS-I \\
$k$ & Index of the intermediate layer for DGS-L \\
$\mathbb{D}^{\left(i\right)}$ & Internal output of $\mathbb{D}$ at layer $i$ \\
$\widetilde{D^{(k)}(S)}$ & Perturbed output of the $k$-th layer in DGS-L\\
$\overline{\mathbb{D}\left(S\right)}$ & Final output from the DGS-L-protected decoder \\
\hline
\hline
\end{tabular}}
\end{table}

\section{Problem Formulation}
In this section, we first present the training process of box-free model watermarking. Then, we introduce an attack-favorable threat model with permissive settings to highlight the effectiveness of our methods. Finally, we outline the detailed steps of our gradient-based watermark removal attack.
\subsection{Box-free Model Watermarking}
\label{sec:basic}
As discussed in Section \ref{sec:introduction} and \ref{sec:boxfree}, box-free model watermarking methods follow a unified framework that jointly trains the watermark embedding network (encoder) $\mathbb{E}$ and the watermark decoding network (decoder) $\mathbb{D}$. Without loss of generality, in this paper, we focus on the post-hoc methods \cite{Zhang2022Deep_Box_Free,Zhang2024Robust_Box_Free} for better illustration. The workflow of box-free model watermarking is depicted in Fig. \ref{fig:victim_architecture} and consists of the following steps
\begin{itemize}
\item Image Generation/Processing: $X \triangleq \mathbb{M}(X_0)$,

\item Watermark Embedding: $Y \triangleq \mathbb{E}({\rm{Concat}}(X, W))$,

\item Watermark Extraction: $\mathbb{D}: \mathcal{Y} \to W$ and $\mathcal{Y}^\complement \to W_0$,
\end{itemize}
where ${\rm{Concat}}(\cdot)$ represents channel-wise concatenation, and $\mathcal{X}_0, \mathcal{X} \subset \mathcal{Y}^\complement$. Keeping $\mathbb{M}$ unchanged, we jointly train $\mathbb{E}$ and $\mathbb{D}$ by minimizing the following objective function
\begin{equation}
    \label{eq:loss_joint}
    \mathcal{L}_{\text{Victim}} = \alpha_1 \mathcal{L}_{\text{Embed}} + \alpha_2 \mathcal{L}_{\text{Fidelity}},
\end{equation}
where $\alpha_1$ and $\alpha_2$ are weighting parameters. Specifically, the loss terms are defined as follows
\begin{align}
    \mathcal{L}_{\text{Embed}} & = \sum_{Y \in \mathcal{Y}}\|\mathbb{D} \left(Y \right) - W \|_2^2 +  \sum_{S \in \mathcal{Y}^\complement}\|\mathbb{D} \left(S \right) - W_0 \|_2^2, \label{eq:loss_robust}\\
    \mathcal{L}_{\text{Fidelity}} & = \sum_{Y \in \mathcal{Y}, X \in \mathcal{X}} \|Y - X \|_2^2,  \label{eq:loss_fidelity}
\end{align}
where $S$ denotes an arbitrary image. Note that $\mathbb{E}$ is implicitly represented by its output $Y$ in (\ref{eq:loss_robust}) and (\ref{eq:loss_fidelity}), which simplifies our subsequent attack formulation, and $\mathbb{E}$ is kept private by the model owner or defender.
\subsection{Threat Model}
\label{sec:threat model}
We discuss the threat model from the perspectives of the model owner, the defender, and the attacker with rigorous assumptions as follows.

\textbf{Model owner.} The model owner implements box-free watermarking with both $\mathbb{E}$ and $\mathbb{D}$ for ownership verification and model theft tracing. As shown in Fig. \ref{fig:victim_architecture}, a black-box API (meaning a query-based interface that hides internal parameters and architecture) is provided to accept query $X_0$ and returns only $Y$, while both $\mathbb{M}$ and $\mathbb{E}$ remain strictly private. Additionally, the black-box API for $\mathbb{D}$ is also provided for watermark verification queries. 

\textbf{Defender.} The role of the defender can be categorized into two scenarios. \textbf{Scenario 1:} The defender is a security service provider, different from the model owner, who does not own $\mathbb{D}$. In this case, the model owner grants the defender limited trust, meaning the defender has access to all resources of $\mathbb{D}$ but is restricted from modifying the original model due to security concerns. \textbf{Scenario 2:} The model owner and the defender refer to the same individual, who has full control over $\mathbb{D}$.

\textbf{Attacker.} The attacker aims to extract $\mathbb{M}$ with a watermark-removed surrogate model. To achieve this, the attacker first selects a set of inputs, $X_0$, and queries the victim model to generate the corresponding outputs, $Y$, mimicking the behavior of normal users. Next, before training the surrogate model, the attacker modifies the outputs $Y$ to remove the embedded watermark, constrained by image quality preservation. Meanwhile, the attacker can query $\mathbb{D}$ to verify whether an image contains the watermark. We also assume that the attacker can access the gradients backpropagated from $\mathbb{D}$’s output, based on existing adversarial attack methods \cite{ilyas2018black_adversarial, Dong2022Query_adversarial, shi2022query_adversarial}. This assumption is similar to the one on unlimited queries to access gradient estimates that leads to the convergence and it provides an attacker-friendly setting to validate the robustness of our proposed defense under the most challenging conditions. Moreover, if the watermark is not removed despite the attacker’s efforts to minimize the removal loss, the attacker could reasonably conclude that defender-imposed gradient perturbations are present and adjust the gradients accordingly as a countermeasure.

\subsection{Gradient-based Removal Attack}
We now present the gradient-based attack for removing box-free watermarks, as illustrated in Fig. \ref{fig:gradient_based_attack}. The core insight behind this approach is similar to the classical adaptive filter, where a feedback loop is used to perform model inversion \cite{Haykin2002Adaptive}, and it comes with a theoretical guarantee of convergence. The attacker can train an inverse of $\mathbb{E}$, denoted as an image-to-image network $\mathbb{R}$, which takes the watermarked image $Y$ and reverses the watermark embedding process. This can be accomplished by using $\mathbb{D}$ as a watermark verifier and an all-white null watermark $W_0$ as the supervision signal, resulting in a minimization problem with the following loss
\begin{equation}
\label{eq:loss_attacker}
\mathcal{L}_\text{Attack} = \beta_1 \mathcal{L}_\text{Removal} + \beta_2 \mathcal{L}_\text{Fidelity}^{\text{A}},
\end{equation}
where $\beta_1$ and $\beta_2$ are the weighting parameters,
\begin{align}
\mathcal{L}_\text{Removal} & =  \|\mathbb{D}[\mathbb{R}(Y)] - W_0\|_2^2, \label{eq:r_loss}\\
\mathcal{L}_\text{Fidelity}^{\text{A}} & = \|\mathbb{R}(Y) - Y\|_2^2 \label{eq:a_fidelity_loss}. 
\end{align}
The removal loss ensures that the modified image $\mathbb{R}(Y)$ does not contain the watermark, while the fidelity loss maintains the quality of the image to ensure its usability for future surrogate training. To minimize (\ref{eq:r_loss}), the gradient is computed as
\begin{align}\label{eq:attack_gradient}
&\nabla {\mathcal{L}_{{\text{Removal}}}} \propto \notag\\ &\frac{{\partial {\mathcal{L}_{{\text{Removal}}}}}}{{\partial \mathbb{D}^{\left(n\right)}\left[ {\mathbb{R}(Y)} \right]}}\left ( \prod_{i=1}^{n-1}\frac{\partial \mathbb{D}^{\left(i+1\right)}\left[ {\mathbb{R}(Y)} \right]}{\partial \mathbb{D}^{\left(i\right)}\left[ {\mathbb{R}(Y)} \right]}   \right ) \frac{{\partial \mathbb{D}^{\left(1\right)}\left[ {\mathbb{R}(Y)} \right]}}{{\partial \mathbb{R}(Y)}},
\end{align}
where $\nabla {\mathcal{L}_{{\text{Removal}}}}$ represents the gradient component used to update the parameters of $\mathbb{R}$ and $\mathbb{D}^{\left(i\right)}\left[ {\mathbb{R}(Y)} \right]$ means the internal output of $\mathbb{D}$ at layer $i$, and $n$ is the total number of layers of $\mathbb{D}$. Although we use the $\ell_2$-norm loss function in (\ref{eq:r_loss}), alternative loss functions could also be employed. Based on our threat model, the gradient of $\mathbb{D}$ with respect to $\mathbb{R}(Y)$, specifically ${\partial \mathbb{D}^{\left(n\right)}\left[ {\mathbb{R}(Y)} \right]}/{\partial \mathbb{R}(Y)}$, is observable, and the attacker has knowledge of all other gradient components, which allows the attacker to effectively train $\mathbb{R}$ in the black-box setting.

\section{Proposed Method}
To defend against the aforementioned gradient-based watermark removal attack, it is unfortunately infeasible to analyze the querying behavior because $\mathbb{D}$, as the watermark extractor, is supposed to accept all types of queries. To solve this problem, an alternative way is to prevent the gradient component in (\ref{eq:attack_gradient}) from being obtained by an attacker, and thus we propose to generate perturbations to (\ref{eq:attack_gradient}) in the black-box querying processing of $\mathbb{D}$, constrained by not altering the watermark extraction performance. Naturally, the perturbation can be added to $\mathbb{D}$'s output, input, or internal layers, which respectively leads to the proposed DGS-O, DGS-I and DGS-L. DGSs, especially DGS-O, is similar to API poisoning based defense for black-box (backdoor) watermarking \cite{Zhang2023Categorical}, but it is formulated herein for box-free watermarking. DGS-O and DGS-I modify output and input of $\mathbb{D}$, respectively, without requiring access to its internal information, so that they can be applied in both \textbf{Scenario 1} and \textbf{Scenario 2} (see Section \ref{sec:threat model}). In contrast, DGS-L operates on the internal layers and requires modifications of the intermediate outputs, making it applicable only to \textbf{Scenario 2}. Furthermore, for decoder $\mathbb{D}$, its query is denoted as $S$, which can fall into three possible situations: \mbox{\textbf{Situation 1:}} $S = Y = \mathbb{E}({\rm{Concat}}(\mathbb{M}(X_0), W)) \in \mathcal{Y}$. This indicates that $S$ is a processed, watermarked image obtained from the black-box API of $\mathbb{M}$, corresponding to a benign query for watermark extraction. \textbf{Situation 2:} $S = \mathbb{R}(Y)$. This represents a malicious query intended for a gradient-based watermark removal attack. \textbf{Situation 3:} $S \in \mathcal{Y}^\complement$. This corresponds to the benign query with a non-watermarked image. We note that a successful defense must prevent the attack in Situation 2 while simultaneously preserving the legitimate users' experience for Situation 1 and Situation 3. In the following content, we formally derive the proposed DGSs. The layer symbol $^{(i)}$ is omitted in DGS-O and DGS-I as they are not used therein.
\subsection{DGS-O}
\subsubsection{Closed-Form Solution}
A defender's primary objective is to prevent watermark removal, focusing on counteracting $\mathcal{L}_\text{Removal}$ as defined in (\ref{eq:r_loss}), while disregarding $\mathcal{L}_\text{Fidelity}^{\text{A}}$ in (\ref{eq:a_fidelity_loss}).
For the ease of presentation, we define $Z \triangleq \mathbb{D}[\mathbb{R}(Y)]$ (see Fig. \ref{fig:gradient_based_attack}) and $Z^{\ast}$ is the modified output returned to the user. Based on (\ref{eq:r_loss}) and (\ref{eq:attack_gradient}), we have
\begin{equation}
\nabla {\mathcal{L}_{{\text{Removal}}}} \propto  \frac{{\partial {\mathcal{L}_{\text{Removal}}}}}{{\partial \mathbb{R}(Y)}} = 2{\left( Z - {W_0} \right)^T}\frac{\partial Z}{{\partial \mathbb{R}(Y)}},\label{eq:original_gradient}
\end{equation}
where $\{\cdot\}^T$ denotes the transpose operator. Training $\mathbb{R}$ requires constructing a dataset of $Y$ by querying the encapsulated $\mathbb{M}$ and $\mathbb{E}$ with a given set of $X_0$. At the initial stage, $\mathbb{R}$ lacks the capability to remove the watermark, leading to $Z \approx W$. To prevent $Z$ from converging to $W_0$, we modify the decoder's output $Z$ as $Z^\ast$ when the decoder's output is closer to the true watermark, i.e., $Z \approx W$, thereby protecting the true gradient direction. Based on (\ref{eq:original_gradient}), the perturbed gradient component is given by 
\begin{equation}
\left(\frac{\partial {\mathcal{L}_{\text{Removal}}}}{\partial \mathbb{R}(Y)}\right)^\ast = 2{\left( {Z^\ast - {W_0}} \right)^T}\frac{{\partial Z^\ast}}{{\partial \mathbb{R}(Y)}}, \label{eq:poison_gradient}
\end{equation}
where the reorientation is designed to ensure that the direction change falls within the range of $90$ to $180$ degrees, i.e.,
\begin{align}
& {\left( {Z^\ast - {W_0}} \right)^T}\frac{{\partial {Z^\ast}}}{{\partial \mathbb{R}(Y)}} = - {\left( {Z - {W_0}} \right)^T}P\frac{{\partial Z}}{{\partial \mathbb{R}(Y)}} \notag\\
\Leftarrow & {\left( {Z^\ast - {W_0}} \right)^T}\frac{{\partial Z^\ast}}{{\partial Z}}= - {\left( {Z - {W_0}} \right)^T}P,\label{eq:z_inverse}
\end{align}
where the chain rule is applied to derive $\partial Z / \partial \mathbb{R}(Y)$ and $P$ is a positive definite matrix. Notably, when $P$ is set to the identity matrix $I$, the reorientation reduces to a simple gradient sign flipping (i.e., a $180$-degree shift). Meanwhile, maintaining the quality of the output image requires that
\begin{equation}
Z^\ast \approx W. \label{eq:z_quality}
\end{equation}
Since (\ref{eq:z_inverse}) represents a first-order differential equation, the solution for $Z^\ast$ takes the following form  
\begin{equation}
Z^\ast = -P Z + C,\label{eq:z_form}
\end{equation}
where $C$ remains independent of $Z$. Substituting (\ref{eq:z_form}) into (\ref{eq:z_quality}) yields
\begin{equation}
-P Z + C \approx W \Rightarrow C \approx (P+I)W, \label{eq:eq}
\end{equation}
which, when substituted back into (\ref{eq:z_form}), leads to
\begin{equation}
\label{eq:ultimate_trans}
Z^\ast = -PZ + (P+I)W,
\end{equation}
where the approximation is replaced by equality for practical implementation. This gradient reorientation mechanism, as described in (\ref{eq:ultimate_trans}), serves as the core component of the proposed DGS-O.

Next, we examine how DGS-O operates across three situations. For \textbf{Situation 1}, as discussed in Section \ref{sec:basic}, the system returns $\mathbb{D}(S) = \mathbb{D}(Y) \approx W$. In \textbf{Situation 2}, since the true gradient is not exposed for $\mathbb{R}$ to facilitate watermark removal, the initial malicious query satisfies $S=\mathbb{R}(Y)\in\mathcal{Y}$. Following the previous discussion, this results in $\mathbb{D}(S) = \mathbb{D}[\mathbb{R}(Y)] = Z \approx W$. For \textbf{Situation 3}, it follows from Section \ref{sec:basic} that $\mathbb{D} (\mathcal{Y}^\complement) \approx W_0$ is returned. Based on these observations and incorporating (\ref{eq:ultimate_trans}), we summarize the response mechanism of DGS-O-protected $\mathbb{D}$ API, denoted as $\mathbb{D}^\ast(S)$, as follows 
\begin{align}\label{eq:final_mechanism}
& \mathbb{D}^\ast(S) = \notag\\
& {{\left\{ {\begin{array}{*{20}{l}}
  { - P\mathbb{D}(S) + (P + I)W,} & {{\text{if }} {\rm{NC}}(\mathbb{D}(S), W) > 0.96}, \\ 
  {\mathbb{D}(S),}&{{\text{otherwise,}}} 
  \end{array}} \right.}}
\end{align}
where ${\rm{NC}(\cdot, \cdot)}$ represents the normalized cross-correlation function and a threshold of $0.96$ is used to determine whether $\mathbb{D}(S) \approx W$, following \cite{Zhang2022Deep_Box_Free}. The above mechanism ensures that for \textbf{Situations 1} and \textbf{2}, which are indistinguishable, the API returns a gradient-reoriented output. In contrast, for \mbox{\textbf{Situation 3}}, it simply provides the original output. Notably, the gradient reorientation in (\ref{eq:final_mechanism}) introduces only imperceptible perturbations to the intended extracted watermark $W$, which preserves the integrity for verification, so its impact on legitimate users is negligible. However, this slight perturbation significantly impedes an attacker’s ability to train $\mathbb{R}$ for watermark removal.

It is important to highlight that, compared to previous gradient-based defenses, such as prediction poisoning \cite{pp_2020_iclr_defense_model_stealing} and gradient redirection \cite{Mantas2022How}, which rely on recursive strategies in protecting deep classification models, (\ref{eq:final_mechanism}) offers a neat closed-form solution to safeguarding the decoder $\mathbb{D}$ in box-free watermarking, effectively mitigating black-box model extraction threats. A key factor in DGS-O is the choice of the positive definite matrix $P$, which determines both the rotation angle and magnitude scaling of the reoriented gradients, directly impacting the overall defense efficacy.
\subsubsection{\texorpdfstring{Choice of $P$}{Choice of P}}
\label{sec:p_choice}
The positive definite matrix $P$ introduced in (\ref{eq:z_inverse}) plays a crucial role in the proposed DGS-O reorientation process. If $P$ is omitted, or equivalently set to $P=I$, an attacker could easily flip the gradient sign and recover the true gradient. In the case when $P \neq I$, consider the eigendecomposition $P = Q^T \Lambda Q$, where $\Lambda$ is a diagonal matrix of eigenvalues and $Q$ is the orthonormal matrix of eigenvectors. When a vector is multiplied on the right of $P$, the process involves first rotating the vector via $Q$, followed by scaling the vector components by the eigenvalues in $\Lambda$, and finally reverting the rotation via $Q^T$. While the rotation introduced by $Q$ is compensated in this process, the indirect rotation caused by scaling with $\Lambda$ is not. For instance, as just one use case, we consider $Q=I$ and thus $P=\Lambda$ is a diagonal matrix with all-positive elements. If these elements differ, applying $-P$ in (\ref{eq:z_inverse}) can result in a rotation of between $90$ to $180$ degrees. It is important to note that an attacker can still flip the gradient sign, but this will cause a deviation of $0$ to $90$ degrees in the gradient. To further enhance the defense, we introduce two guidances for practice. \textbf{1) Gradient Norm Reduction:} By setting the diagonal elements of $\Lambda_i$ such that $0 < \Lambda_i \ll 1$, where $\Lambda_i$ is the $i$th diagonal element, we can reduce the gradient norm so that even an attacker succeeds in recovering the true gradient direction, the learning rate will become vanishingly small. This technique also suggests that in the limit when all elements of $P$ are zero, previous hard-decision methods directly return $Z^\ast = W$ to the user is reasonable to some extent. \textbf{2) Randomness Injection:} In \textbf{Situation 1} and \textbf{Situation 2}, we can independently sample the diagonal elements of $P$ from a distribution (e.g., uniform distribution) for each query. This randomization ensures that even if the attacker becomes aware of the defense mechanism, it will be impossible to correct the gradient direction during each training iteration for $\mathbb{R}$. The accumulated bias will further prevent $\mathbb{R}$ from convergence.

\subsubsection{Limitation Analysis}
\label{sec:dgso_limitation_analysis}
We observe that although \mbox{DGS-O} offers a simple yet effective way to protect $\mathbb{D}$, its inherent vulnerability stems from its closed-form nature and fixed applied location. As indicated in (\ref{eq:ultimate_trans}), if an attacker can recover the hidden $Z$ from the perturbed $Z^\ast$, DGS-O can be completely overcome. To achieve this, the inverse transformation of (\ref{eq:ultimate_trans}) is given by
\begin{equation}
    Z = - P^{-1}Z^{\ast} + \left(I + P^{-1}\right)W, \label{eq:limitation}
\end{equation}
which requires knowledge of both $W$ and $P$. While $W$ can be estimated by querying $\mathbb{D}$ with a watermarked image $Y \in \mathcal{Y}$, determining $P$ remains challenging. However, an attacker may approximate $Z$ by assuming $P=I$ and replacing $W$ with $\mathbb{D}^{\ast}(Y)$ in (\ref{eq:limitation}), leading to the estimate $\Tilde{Z} = - Z^{\ast} + 2 \mathbb{D}^{\ast}(Y)$. Although based on our experiments in Section \ref{sec:robustness}, this approximation fails to compromise the defense, it remains an open question whether attackers can develop more advanced strategies to further refine the estimation of $Z$. Next, we present DGS-I and DGS-L that overcome this limitation.

\begin{figure}[!t]
    \centering
    \includegraphics[width=0.6\linewidth]{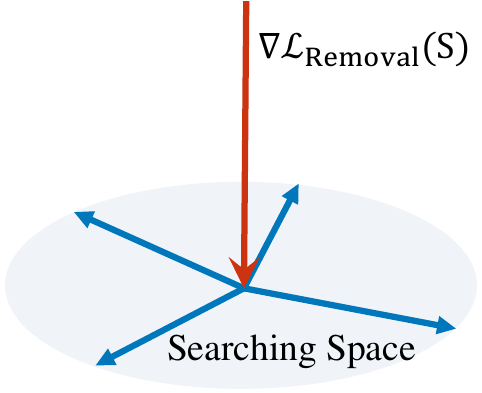}
    \caption{Illustration of proposed orthogonality-based perturbation selection mechanism.}
    \label{fig:orthogonality}
    \vspace{-3mm}
\end{figure}

\subsection{DGS-I}
DGS-I is inspired by the well-known FGSM \cite{goodfellow2014explaining}. It imposes adversarial perturbations on the input to $\mathbb{D}$, reorienting the gradient used to train $\mathbb{R}$ in (\ref{eq:attack_gradient}) while preserving the output image quality. 
We begin by discussing how to generate the perturbation based on prior adversarial research and its limitations. Then, to further enhance the defense, we introduce orthogonality as a guiding mechanism for perturbation selection, ensuring resilience against gradient recovery by the attacker while offering a provable performance guarantee.

\subsubsection{Perturbation Design}
Given an input $S$ to $\mathbb{D}$, the adversarial sample $\tilde{S}$ is defined as
\begin{equation}
\tilde{S} = S + \eta\left(S\right), \quad \text{s.t.} \quad \|\eta\left(S\right)\|_{\infty} \le \epsilon,
\end{equation}
where $\eta\left(S\right)$ is an adversarial perturbation and $\epsilon$ is a small scalar controlling the deviation between $\tilde{S}$ and $S$. The objective of crafting such an adversarial example is to maximize $\mathcal{L}_\text{Removal}$ in (\ref{eq:r_loss}), thereby preventing the convergence of the training algorithm for $\mathbb{R}$. Since $\eta\left(S\right)$ is a small number, we can employ the first-order Taylor expansion on $\mathcal{L}_\text{Removal}$ at $S$ to obtain
\begin{equation}
    \label{eq:taylor_r_loss}
    \mathcal{L}_\text{Removal}\left(\tilde{S}\right) \approx \mathcal{L}_\text{Removal}\left(S\right) + \nabla{\mathcal{L}_\text{Removal}\left(S\right)^T}\eta\left(S\right),
\end{equation}
where $\eta\left(S\right) = \tilde{S} - S$. To maximize it, denote the $i$th element of $\eta\left(S\right)$ by $\eta_i\left(S\right) = \epsilon \text{sign}\left(\nabla{\mathcal{L}_\text{Removal}\left(S\right)_i}\right)$, where $\text{sign}(\cdot)$ is denoted as
\begin{equation}
\text{sign}\left(\nabla{\mathcal{L}_\text{Removal}\left(S\right)_i}\right) = 
\begin{cases} 
-1, & \text{if } \nabla{\mathcal{L}_\text{Removal}\left(S\right)_i} < 0, \\
0, & \text{if } \nabla{\mathcal{L}_\text{Removal}\left(S\right)_i} = 0, \\
1, & \text{if }  \nabla{\mathcal{L}_\text{Removal}\left(S\right)_i} > 0,
\end{cases}
\end{equation}
then, for any query $S$ of $\mathbb{D}$, we can construct 
\begin{equation}
    \label{eq:preliminary_dgsi}
    \tilde{S} = S + \epsilon\text{sign}\left(\nabla{\mathcal{L}_\text{Removal}\left(S\right)}\right),
\end{equation}
as a countermeasure against the gradient-based attack. Considering the attacker's gradient chain in (\ref{eq:attack_gradient}), it follows from the perturbed input that
\begin{align}
&\frac{\partial\mathcal{L}_\text{Removal}}{\partial S} \rightarrow \nabla{\mathcal{L}_\text{Removal}}\left(\tilde{S}\right)^T\left(I + \frac{\partial \eta(S)}{\partial S}\right) \approx \notag\\
 &\left(\nabla{\mathcal{L}_\text{Removal}\left(S\right)}+\nabla^{2}{\mathcal{L}_\text{Removal}\left(S\right)}\eta\left(S\right)\right)^T\left(I + \frac{\partial \eta(S)}{\partial S}\right).
\label{eq:2nd_order_preliminary_dgsi}
\end{align}
The approximation is derived from the first-order Taylor expansion at $S$, leading to extra perturbation terms in the calculation of $\frac{\partial\mathcal{L}_\text{Removal}}{\partial S}$. Subsequently, deviation of gradient direction accumulates per iteration, preventing the convergence of $\mathbb{R}$.

However, considering \textbf{Situation 3} where $S \in \mathcal{Y}^\complement$, i.e., the benign query with a non-watermarked image, the perturbation in (\ref{eq:preliminary_dgsi}) increases the value of  $\|\mathbb{D}(\tilde{S}) - W_0\|^2_2$ and may cause the output to deviate from $W_0$, potentially affecting the normal user experience. Moreover, once the attacker becomes aware of the defense mechanism, they can simply flip the sign of $\mathcal{L}_\text{Removal}$ to restore loss minimization. To address these issues, we introduce a revised perturbation selection method based on orthogonality as follows.

\subsubsection{Orthogonal Perturbation Selection}
Compared with the original $\mathcal{L}_\text{Removal}$, we note that the limitations discussed above arise from the additional term $\nabla{\mathcal{L}_\text{Removal}\left(S\right)^T}\eta\left(S\right)$ in (\ref{eq:taylor_r_loss}). An intuitive idea is to remove this extra term while preserving the perturbation in the attacker's backpropagated gradient. To achieve this, we can randomly choose an adversarial perturbation that satisfies the orthogonality condition $\nabla{\mathcal{L}_\text{Removal}\left(S\right)^T} 
\perp \eta\left(S\right)$ so that $\nabla{\mathcal{L}_\text{Removal}\left(S\right)^T}\eta\left(S\right)=0$. In this case, the attacker's gradient chain in (\ref{eq:attack_gradient}) can be reformulated similarly to (\ref{eq:2nd_order_preliminary_dgsi}) as
\begin{multline}
\label{eq: gradient_expansion_ortho}
    \frac{\partial\mathcal{L}_\text{Removal}}{\partial S} \rightarrow 
    \nabla{\mathcal{L}_\text{Removal}\left(S\right)}^T + \nabla{\mathcal{L}_\text{Removal}\left(S\right)}^T\frac{\partial\eta\left(S\right)}{\partial S} \\
    + \left(\nabla^2{\mathcal{L}_\text{Removal}}\eta\left(S\right)\right)^T +\left(\nabla^2{\mathcal{L}_\text{Removal}\left(S\right)}\eta\left(S\right)\right)^T\frac{\partial\eta\left(S\right)}{\partial S}.
\end{multline}
Differentiating both sides of $\nabla{\mathcal{L}_\text{Removal}\left(S\right)^T}\eta\left(S\right)=0$, we obtain
\begin{equation}
\nabla{\mathcal{L}_\text{Removal}\left(S\right)}^T\frac{\partial\eta\left(S\right)}{\partial S} 
    + \left(\nabla^2{\mathcal{L}_\text{Removal}}\eta\left(S\right)\right)^T = 0,
\end{equation}
and then substituting it into (\ref{eq: gradient_expansion_ortho}), the gradient simplifies to
\begin{equation}
    \nabla{\mathcal{L}_\text{Removal}\left(S\right)}^T + \left(\nabla^2{\mathcal{L}_\text{Removal}\left(S\right)}\eta\left(S\right)\right)^T\frac{\partial\eta\left(S\right)}{\partial S},
\end{equation}
leaving behind an interference term in the chain. We can observe that this orthogonality-guided perturbation does not affect normal users' experience since it keeps $\|\mathbb{D}(\tilde{S}) - W_0\|^2_2$ intact. Additionally, as illustrated in Fig. \ref{fig:orthogonality}, even if an attacker realizes the existence of the defense mechanism, the orthogonality and randomized nature of our perturbation selection guarantee that the search space for the attacker is infinite, making it nearly impossible to reconstruct the original gradient.

\begin{figure*}[!t]
    \centering
    \includegraphics[width=.98\linewidth]{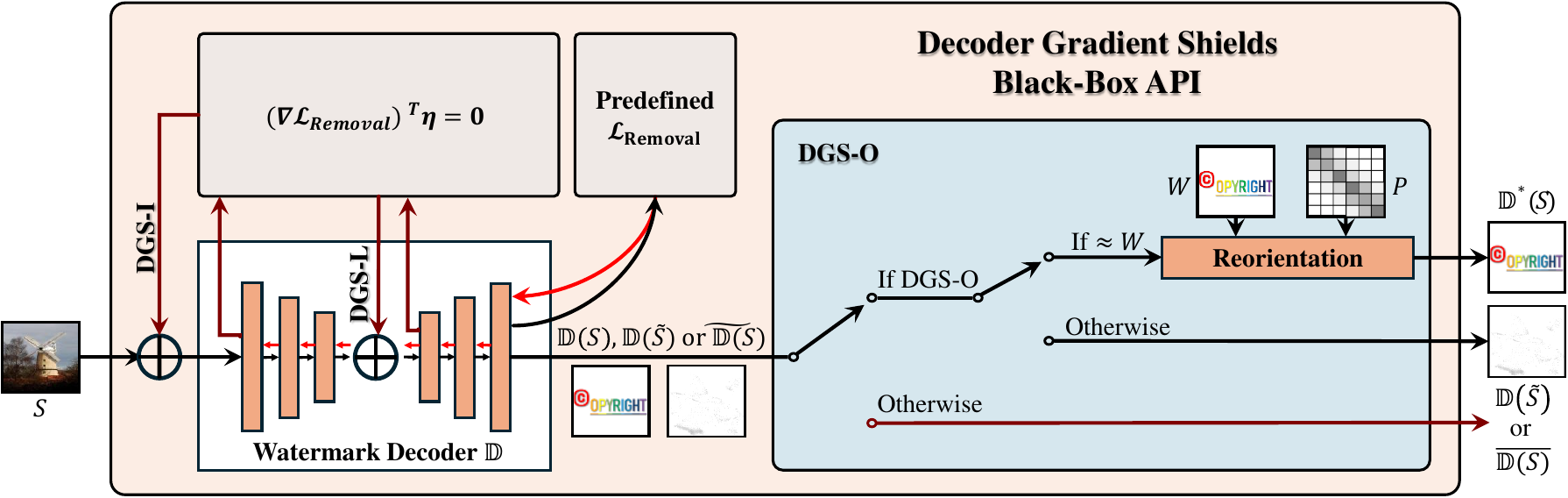}
    \caption{Flowchart of the proposed DGSs in the black-box API of $\mathbb{D}$. The red arrows represent backpropagated gradients.}
    \label{fig:dgs_framework}
\end{figure*}

\subsection{DGS-L}
Although DGS-I overcomes the primary drawbacks of DGS-O, it still has some limitations: 1) It also employs a fixed manner that the perturbation is applied at the input, which lacks flexibility and may allow an attacker to easily infer the location of the defense, thereby facilitating further attacks such as a dedicated modification on $\mathbb{D}$'s input to bypass the defense. 2) DGS-I requires computing the full backpropagated gradient within $\mathbb{D}$, making the perturbation generation process computationally expensive. To overcome these limitations, we introduce the third method in our DGSs family, DGS-L. It retains the perturbation generation approach in DGS-I but applies it to the output of a certain internal layer of $\mathbb{D}$, making this method more flexible and efficient than DGS-I.

Let the perturbed layer number be $k$, where $k \in [1,n-1]$, whose perturbed output is denoted by $\widetilde{D^{(k)}(S)}$, then DGS-L turns the internal gradient chain with respect to input $S$ in (\ref{eq:attack_gradient}) into
\begin{equation}
   \left ( \prod_{i=k}^{n-1}\frac{\partial \mathbb{D}^{\left(i+1\right)}\left(S\right)}{\partial \widetilde{\mathbb{D}^{\left(i\right)}\left(S \right)}}\right )\frac{\partial \widetilde{\mathbb{D}^{\left(k\right)}\left(S \right)}}{\partial \mathbb{D}^{\left(k\right)}\left(S \right)}\left ( \prod_{i=1}^{k-1}\frac{\partial \mathbb{D}^{\left(i+1\right)}\left(S\right)}{\partial \mathbb{D}^{\left(i\right)}\left(S \right)}\right). 
\end{equation}
DGS-L splits $\mathbb{D}$ into two separate components (denoted as preceding and subsequent component) and adds perturbations generated by the orthogonality-based mechanism to the input of the subsequent component (i.e., the output of the preceding component) to interfere with an attacker's backpropagated gradients. Therefore, the derivation process is similar to that of DGS-I, which is omitted for brevity. DGS-L can reduce the computational overhead of generating adversarial perturbations due to fewer gradients to compute, conceal the location where perturbations are applied, and maintain provable performance. However, it requires access to the intermediate-layer outputs of $\mathbb{D}$, meaning it can only be deployed when the defender has access to the decoder's internal information, i.e., under \textbf{Scenario 2}. We denote the output of $\mathbb{D}$ when DGS-L is applied as $\overline{\mathbb{D}\left(S\right)}$.

\subsection{Summary of DGSs}
\label{sec:summary_dgss}
The flowchart of the proposed DGSs, including DGS-O, DGS-I, and DGS-L, is illustrated in Fig. \ref{fig:dgs_framework}. Below, we provide a brief summary of each method.
\begin{itemize}
    \item \textbf{DGS-O:} DGS-O offers a closed-form solution that reorients an attacker's gradient. It is well-suited for time-sensitive defense and is applicable to both \textbf{Scenario 1} ($\text{defender} \neq \text{model owner}$) and \textbf{Scenario 2} ($\text{defender} = \text{model owner}$).
    \item \textbf{DGS-I:} DGS-I mitigates the theoretical vulnerability of DGS-O by introducing an orthogonality-guided adversarial perturbation at $\mathbb{D}$'s input. However, it requires the computation of the full gradient chain. DGS-I is suitable for defenses where time constraints are less stringent but security demands are higher, and it is also applicable to both \textbf{Scenario 1} and \textbf{Scenario 2}.
    \item  \textbf{DGS-L:} DGS-L enhances DGS-I by improving both security and computational efficiency. However, it is only applicable to \textbf{Scenario 2}, where the defender can modify the layer outputs in $\mathbb{D}$.
\end{itemize}
We further claim that simultaneously employing multiple DGS methods is unnecessary, as all the DGSs result in disturbing an attacker's gradient. Instead, it would introduce additional computational overhead and potential conflicts, which could negatively impact the normal user experience.

\begin{figure}[!t]
    \centering
    \includegraphics[width=.95\linewidth]{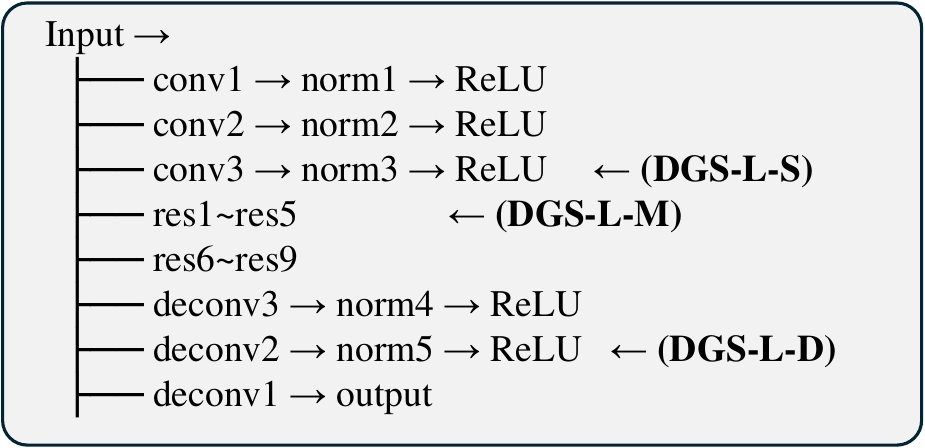}
    \caption{Illustration of the three DGS-L variants applied to $\mathbb{D}$, where ``S'', ``M'', and ``D'' denote the shallow, middle, and deep layer implementations, respectively}
    \label{fig:DGS-L}
\end{figure}

\section{Experimental Results}
In this section, we present comprehensive experimental results to validate the effectiveness of the proposed DGSs. The evaluations are conducted on the state-of-the-art box-free model watermarking framework \cite{Zhang2024Robust_Box_Free}. We note that \cite{Zhang2024Robust_Box_Free} is an extended version of \cite{Zhang2022Deep_Box_Free}, which addresses the vulnerabilities of the original watermarking scheme to image augmentation attacks. Nevertheless, since both works share the same watermark encoder and decoder, and this paper does not consider the impact of image augmentation, we treat them as the same framework to avoid redundant comparisons.

\subsection{Experimental Settings}
\subsubsection{Datasets}
We consider two representative tasks: image deraining (a classical low-level image processing problem) and text-to-image (T2I) generation (a high-level computer vision task). For the deraining task, we adopt the PASCAL VOC dataset \cite{everingham2010pascal_Dataset_Pascal}, where its data corresponds to $X$ in Fig.~\ref{fig:victim_architecture}. The $12,000$ training images are evenly divided into two subsets of $6,000$ images each, used for training the victim model and the remover, respectively. The rainy images, corresponding to $X_0$ in Fig.~\ref{fig:victim_architecture}, are synthesized using the algorithm from \cite{zhang2018density}. For the image generation task, a separate set of $12,000$ text prompts is also split evenly for training the victim model and the remover. The text prompts serve as $X_0$, and the corresponding images $X$ are generated using the famous Stable Diffusion \cite{rombach2022high_Victim_StableDiffusion}. Additionally, all images in both tasks are in RGB format with a resolution of $256 \times 256$.
\subsubsection{Metrics}
For fidelity evaluation, we adopt the peak signal-to-noise ratio (PSNR) and the multi-scale structural similarity index (MS-SSIM) \cite{wang2003multiscale} to assess the similarity between the original and output images. For robustness evaluation, we report the defense success rate, denoted as $\text{SR}$, which is defined as the ratio of watermarked images from which the embedded watermark can be successfully extracted after undergoing removal attacks.
\subsubsection{Implementation Details}
We adopt the same model structure, training procedure, and  hyperparameter settings as the victim model \cite{Zhang2024Robust_Box_Free}, incorporating $\mathbb{M}$, $\mathbb{E}$, and $\mathbb{D}$. The gradient-based remover $\mathbb{R}$ is constructed using UNet \cite{ronneberger2015u_UNet}. Both models are trained from scratch for $100$ epochs using the Adam optimizer with a fixed learning rate of $0.0002$. The weighting coefficients $\alpha_1$, $\alpha_2$, $\beta_1$, and $\beta_2$ in Equations (\ref{eq:loss_joint}) and (\ref{eq:loss_attacker}) are all set to 1. For DGS-L, we explore three locations for perturbation, as depicted in Fig. \ref{fig:DGS-L}, where ``S'', ``M'', and ``D'' indicate DGS-L implementation at $\mathbb{D}$'s shallow, middle, and deep layers, respectively.

\begin{figure*}[!t]
    \centering

    \begin{minipage}[c]{0.04\linewidth} 
        \centering
        \rotatebox{90}{\small DGS-O (Deraining)} 
    \end{minipage}
    \begin{minipage}[c]{0.96\linewidth} 
        \subfloat[$\ell_1$ loss]{%
            \includegraphics[width=0.33\linewidth]{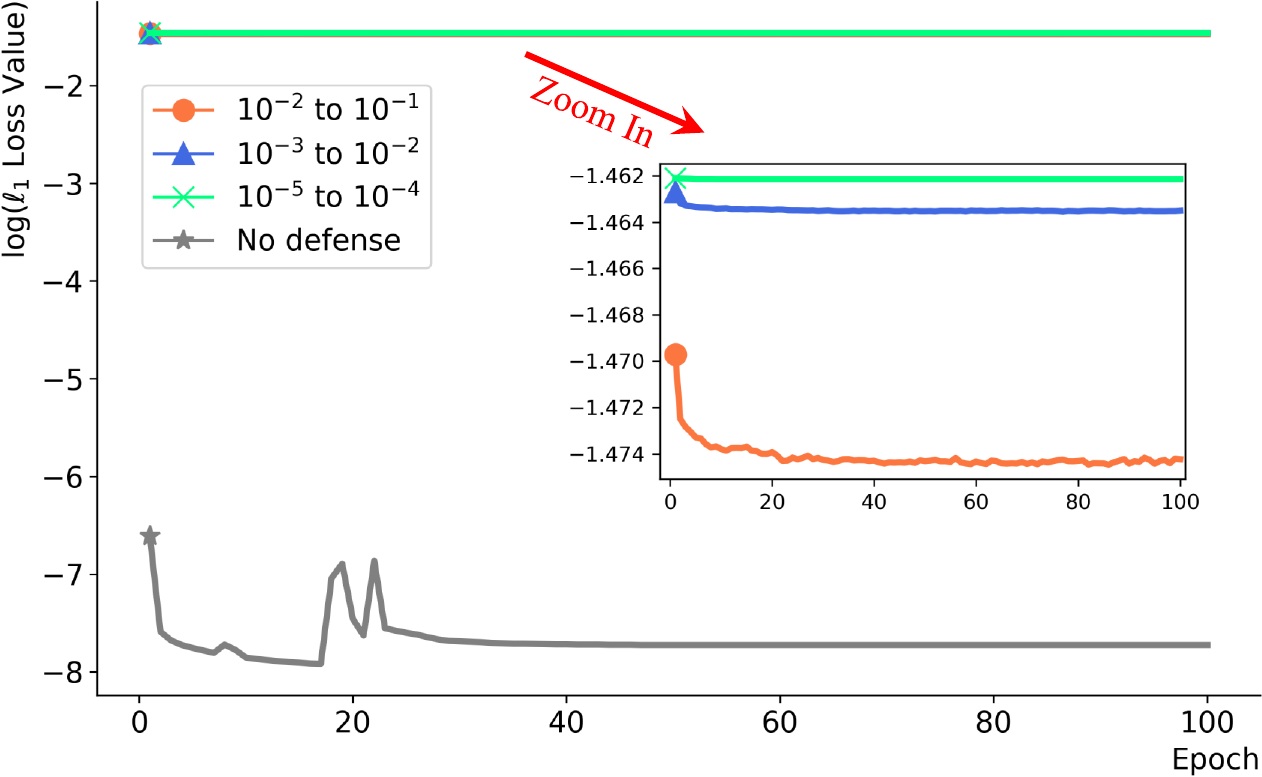}%
        }
        \hfill
        \subfloat[$\ell_2$ loss]{%
            \includegraphics[width=0.33\linewidth]{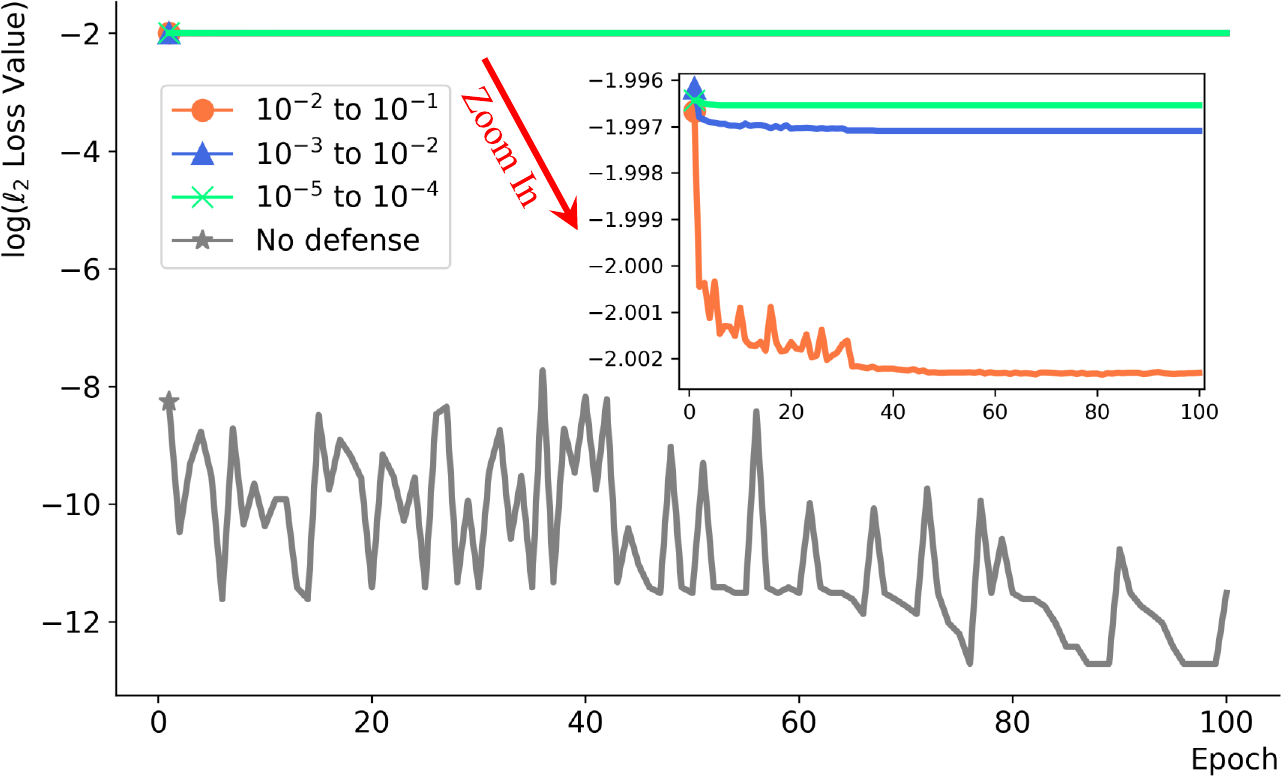}%
        }
        \hfill
        \subfloat[$\ell_2$ and consistent losses]{%
            \includegraphics[width=0.33\linewidth]{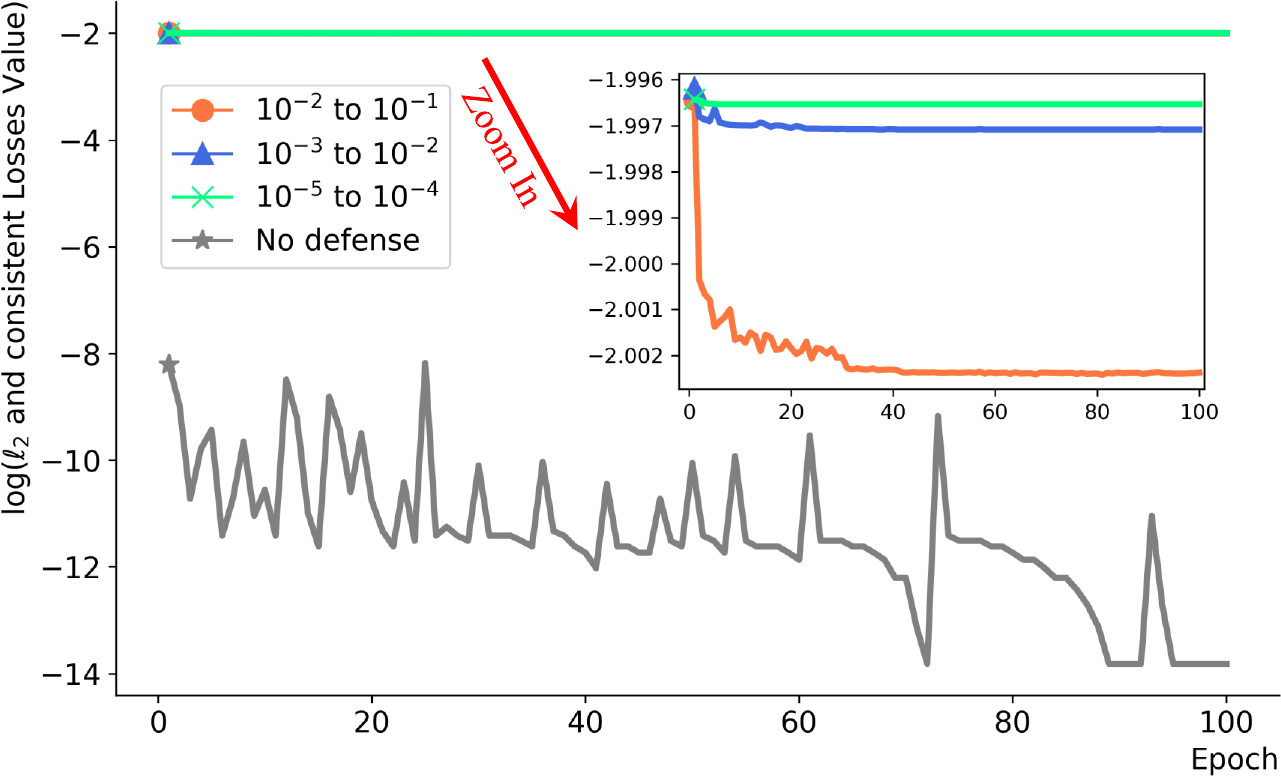}%
        }
    \end{minipage}

    \begin{minipage}[c]{0.04\linewidth} 
        \centering
        \rotatebox{90}{\small DGS-O (Image Gen.)}
    \end{minipage}
    \begin{minipage}[c]{0.96\linewidth} 
        \subfloat[$\ell_1$ loss]{%
            \includegraphics[width=0.33\linewidth]{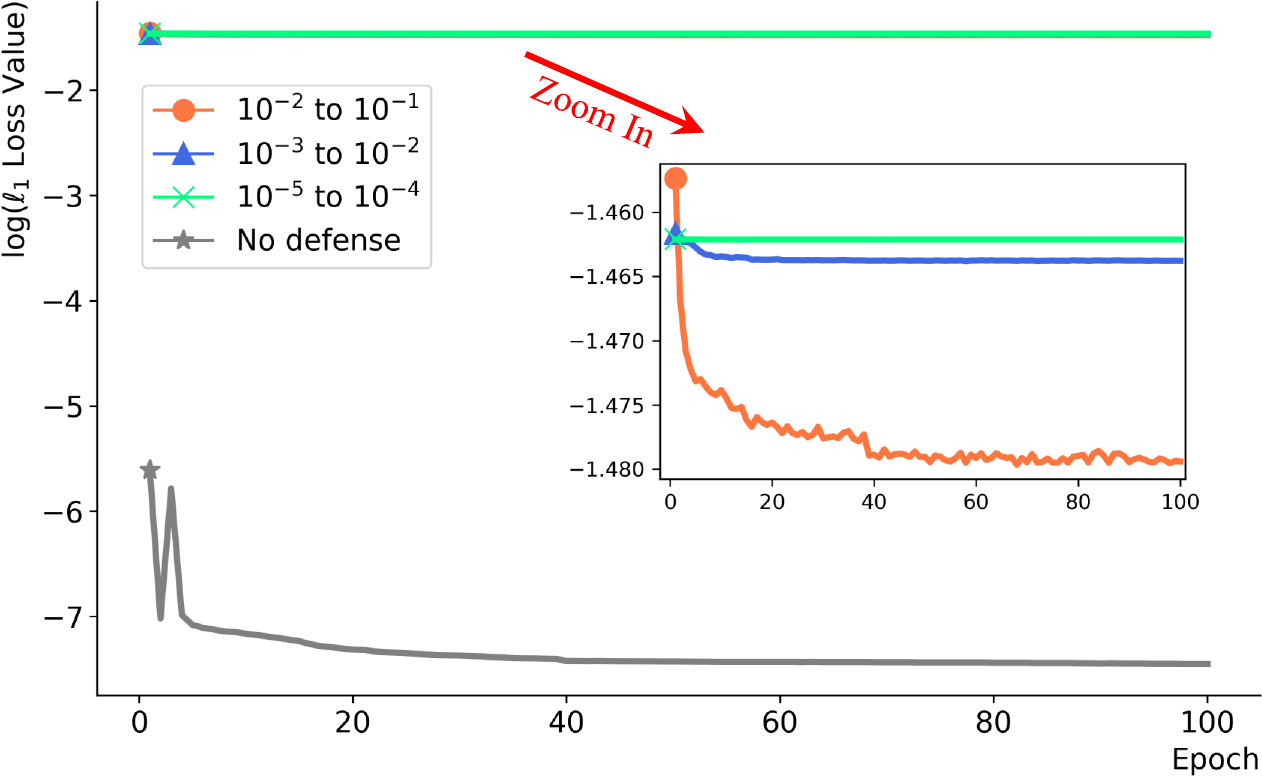}%
        }
        \hfill
        \subfloat[$\ell_2$ loss]{%
            \includegraphics[width=0.33\linewidth]{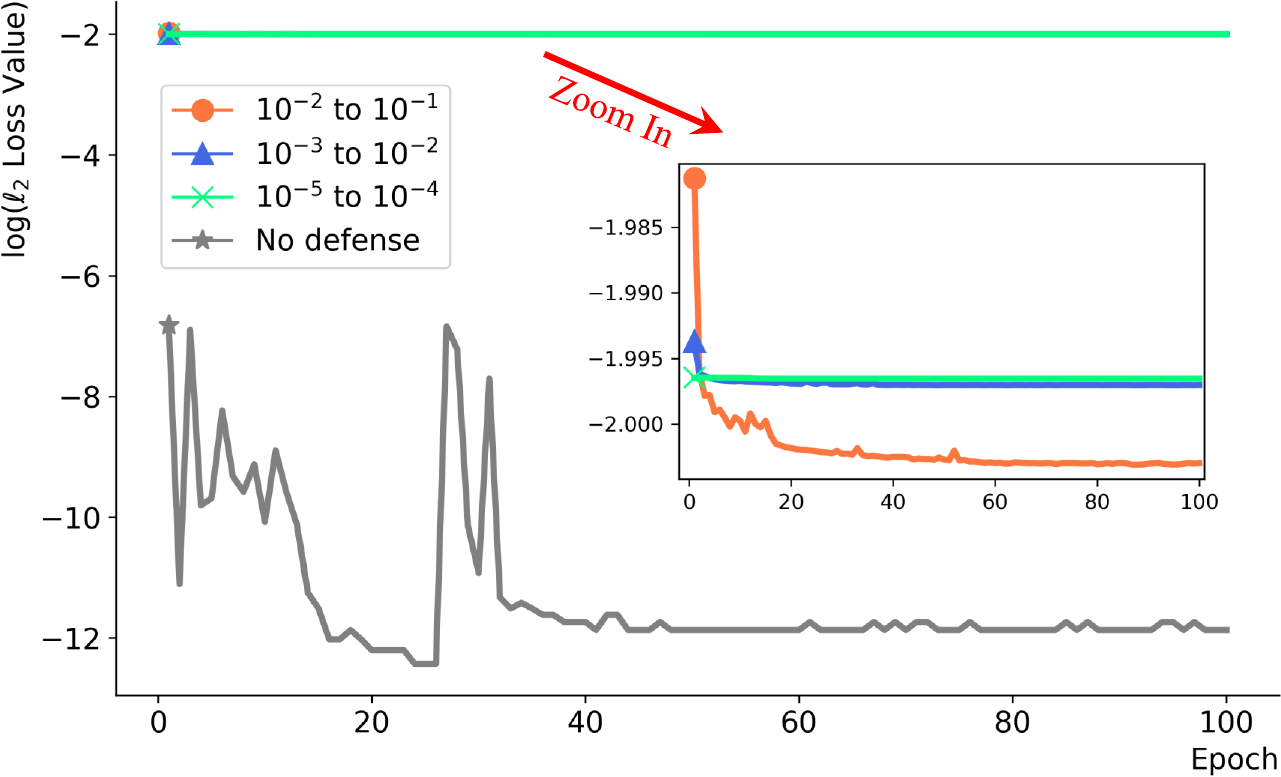}%
        }
        \hfill
        \subfloat[$\ell_2$ and consistent loss]{%
            \includegraphics[width=0.33\linewidth]{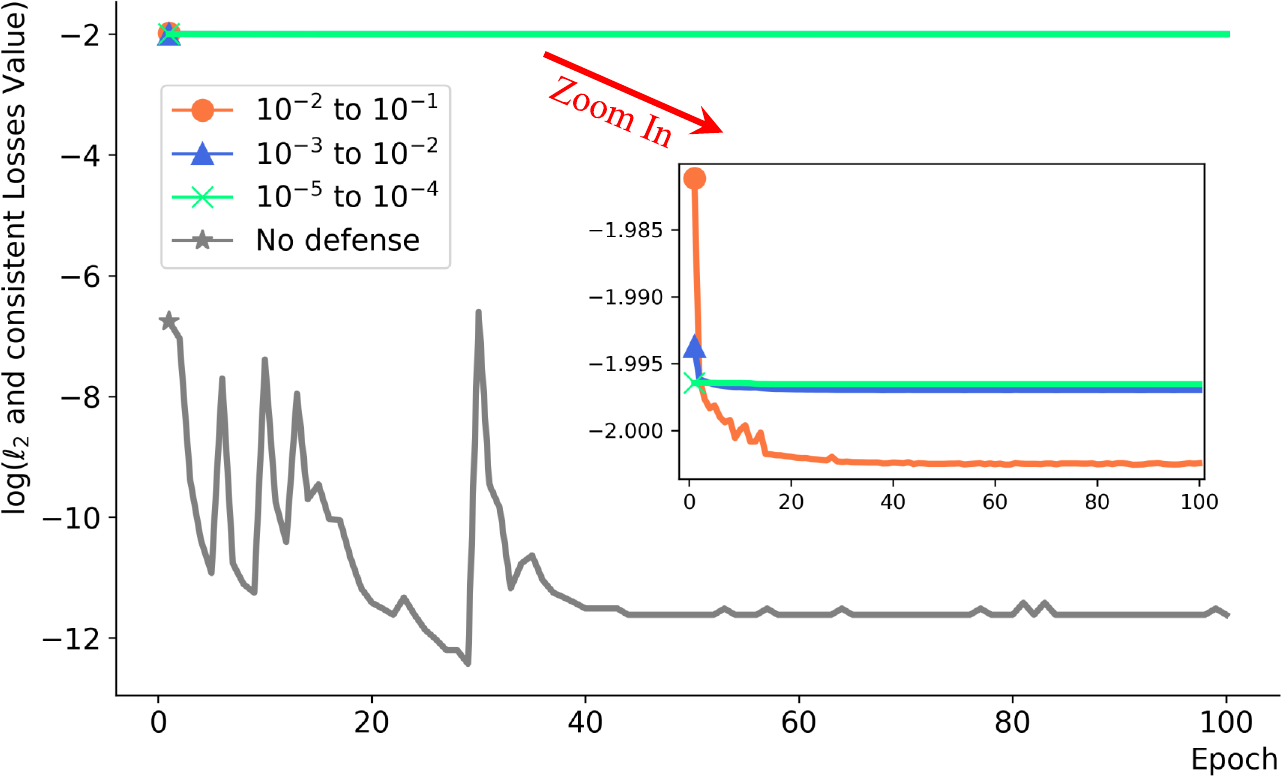}%
        }
    \end{minipage}

    \begin{minipage}[c]{0.04\linewidth} 
        \centering
        \rotatebox{90}{\small DGS-I/L (Deraining)}
    \end{minipage}
    \begin{minipage}[c]{0.96\linewidth} 
        \subfloat[$\ell_1$ loss]{%
            \includegraphics[width=0.33\linewidth]{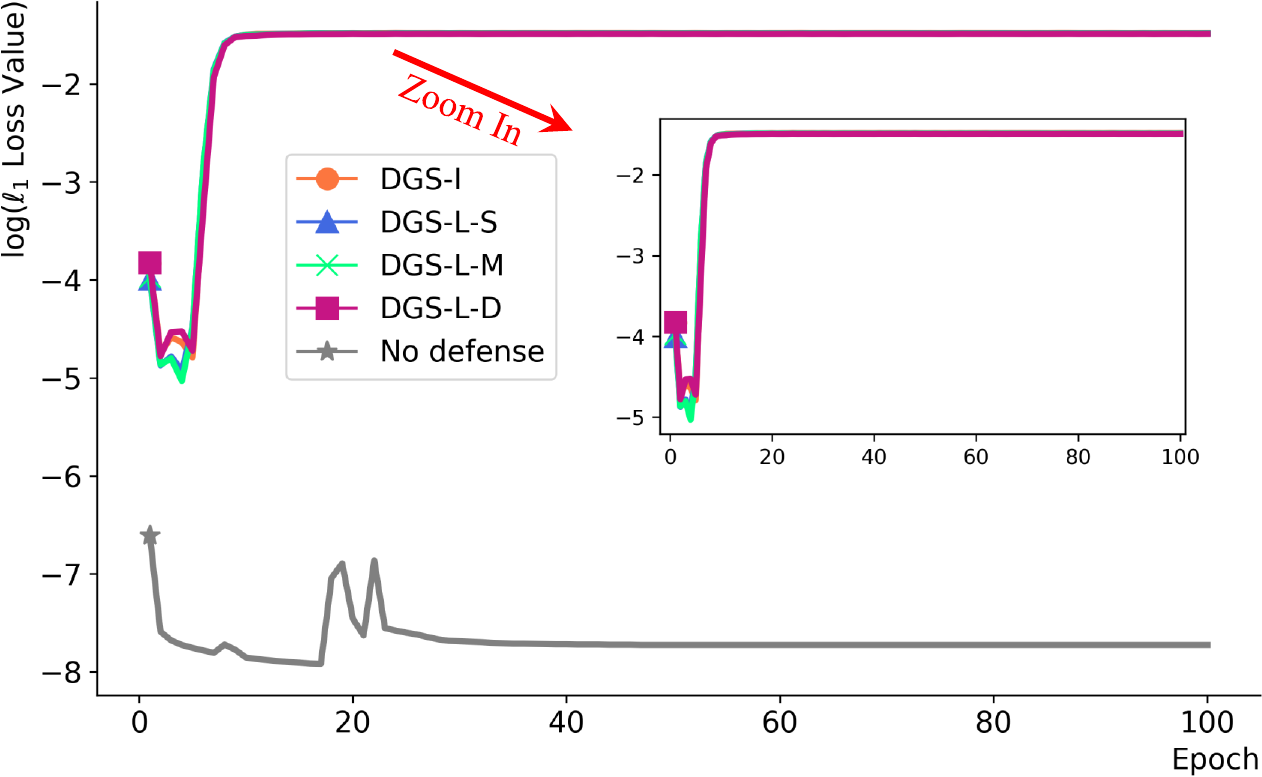}%
        }
        \hfill
        \subfloat[$\ell_2$ loss]{%
            \includegraphics[width=0.33\linewidth]{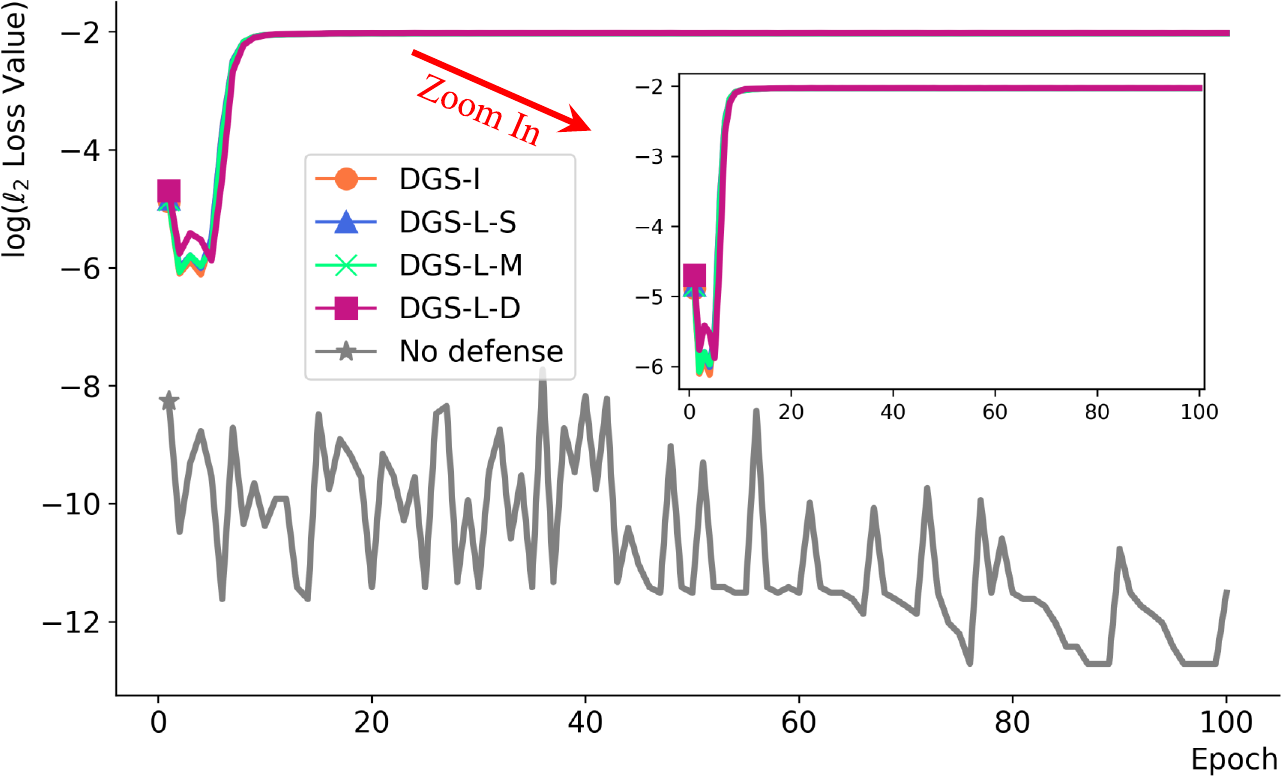}%
        }
        \hfill
        \subfloat[$\ell_2$ and consistent loss]{%
            \includegraphics[width=0.33\linewidth]{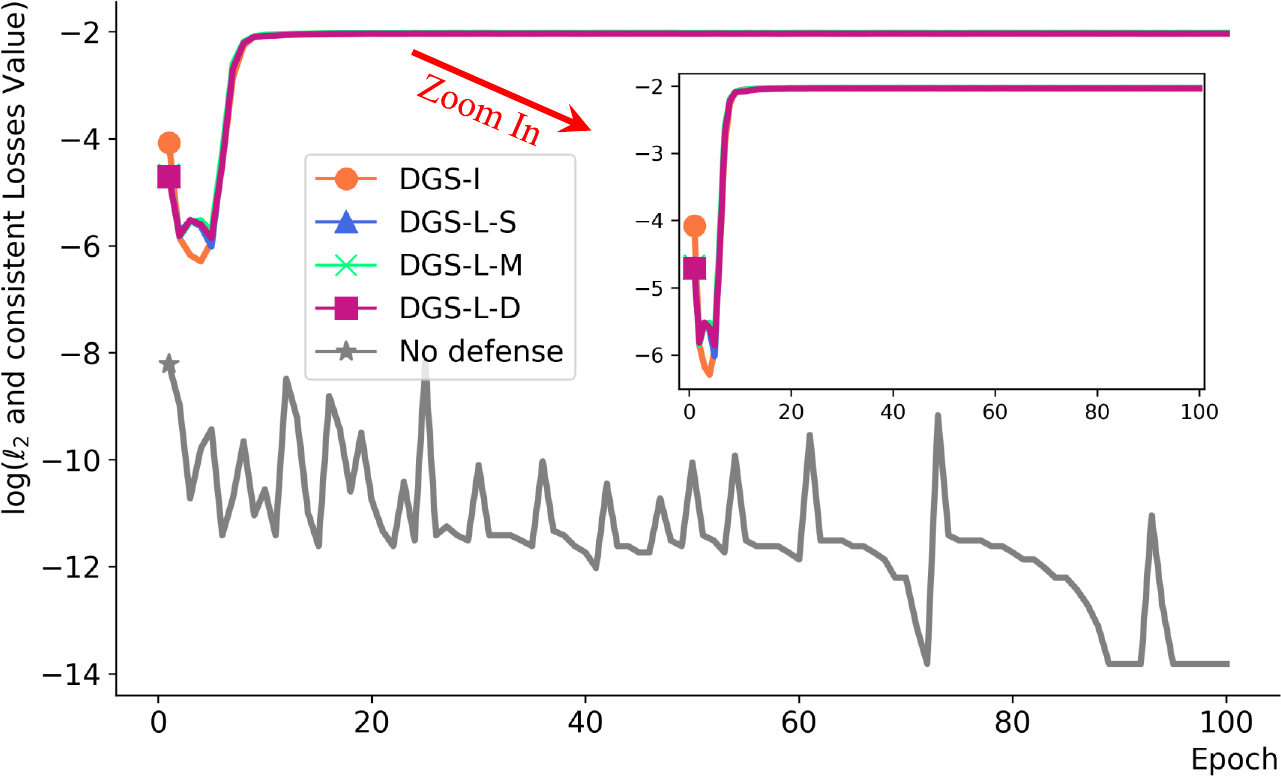}%
        }
    \end{minipage}

    \begin{minipage}[c]{0.04\linewidth} 
        \centering
        \rotatebox{90}{\small DGS-I/L (Image Gen.)}
    \end{minipage}
    \begin{minipage}[c]{0.96\linewidth} 
        \subfloat[$\ell_1$ loss]{%
            \includegraphics[width=0.33\linewidth]{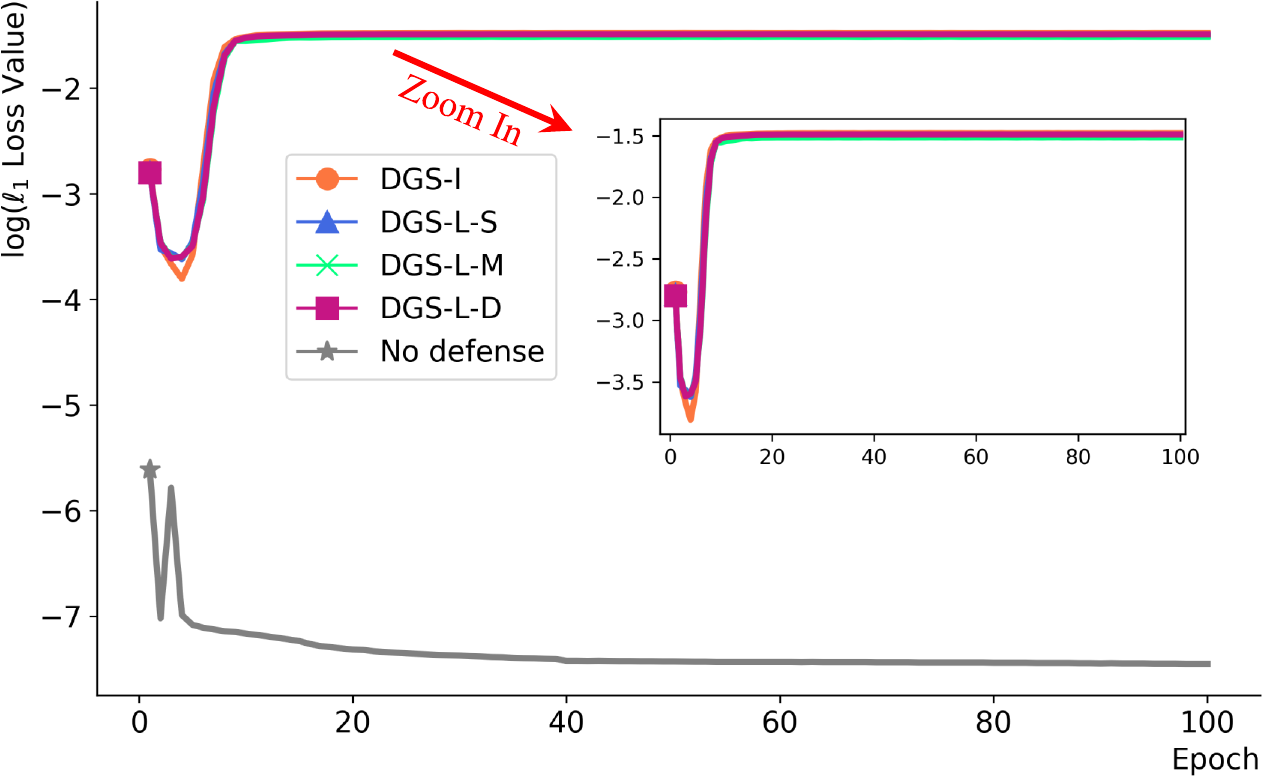}%
        }
        \hfill
        \subfloat[$\ell_2$ loss]{%
            \includegraphics[width=0.33\linewidth]{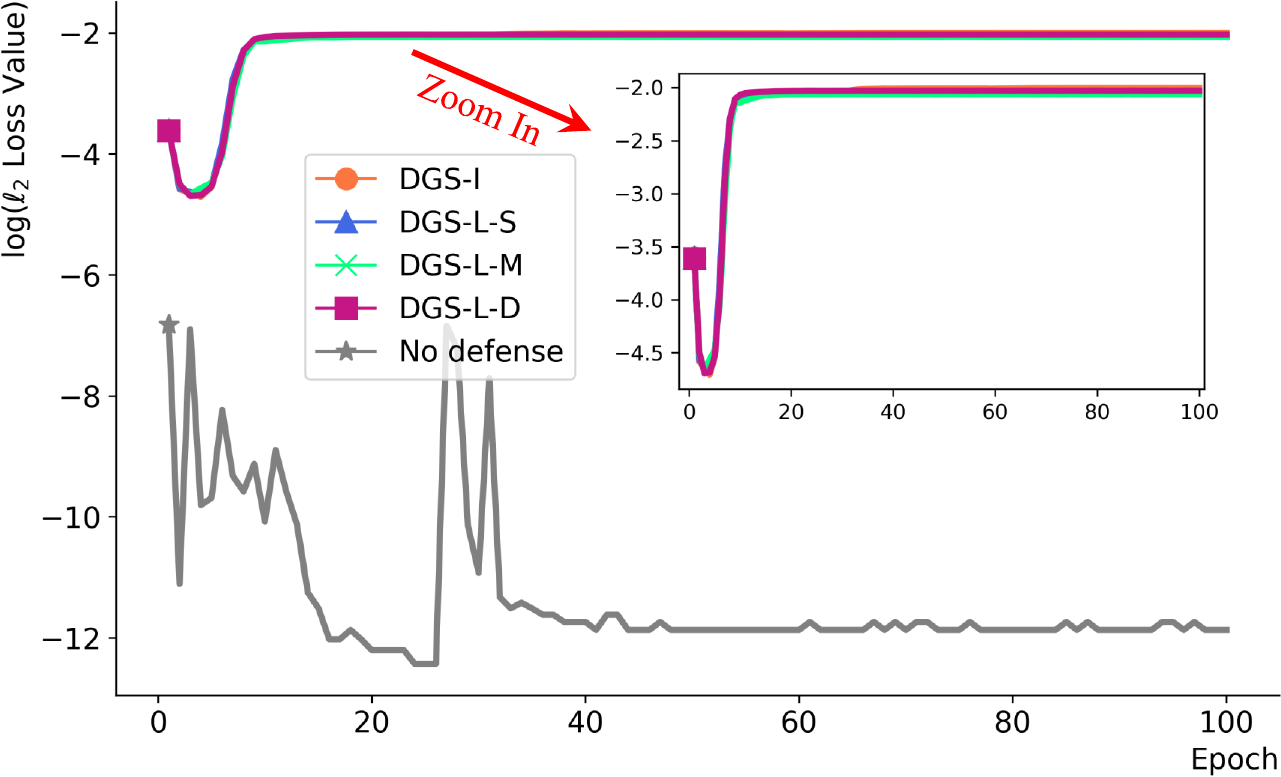}%
        }
        \hfill
        \subfloat[$\ell_2$ and consistent loss]{%
            \includegraphics[width=0.33\linewidth]{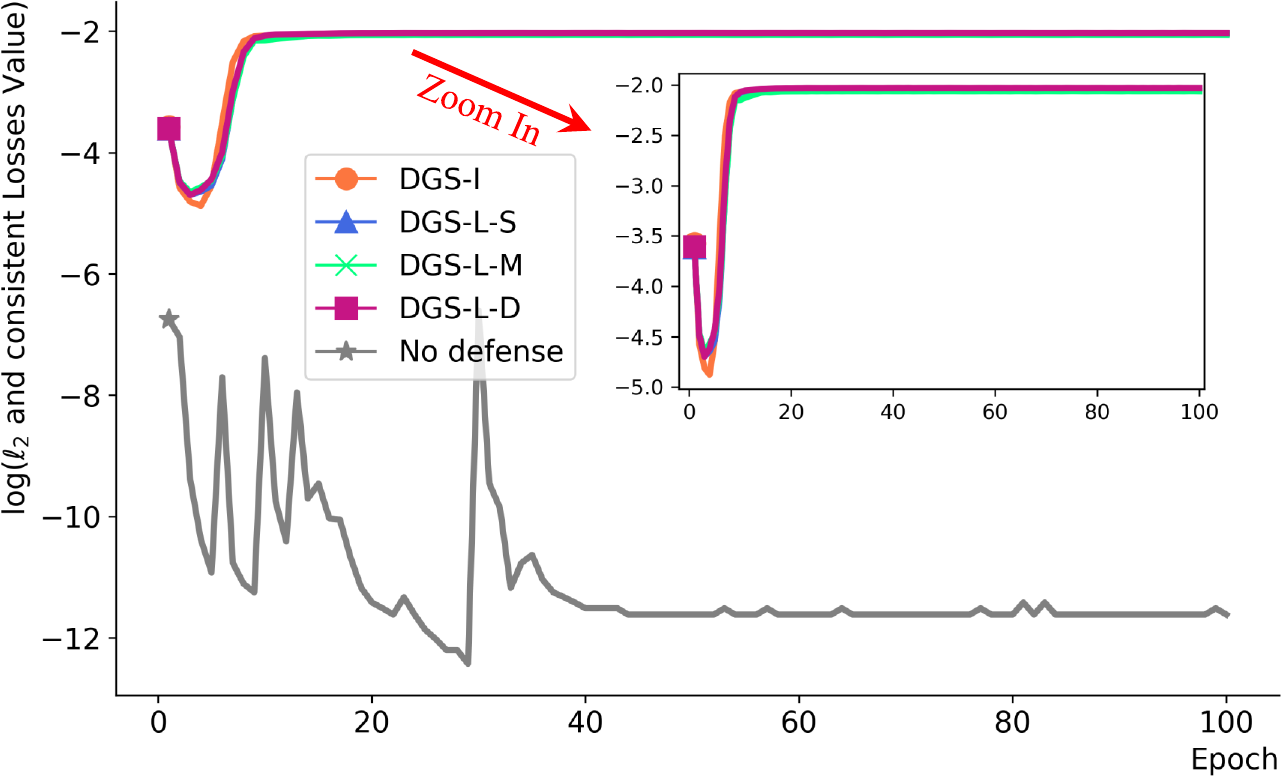}%
        }
    \end{minipage}

    \caption{Quantitative evaluation of the convergence behavior of the attacker's different removal loss functions during the training of $\mathbb{R}$ under different DGS configurations. The first and second rows correspond to DGS-O with different choices of $P$, applied to deraining and image generation tasks, respectively. The third and fourth rows depict the performance of DGS-I and DGS-L on the deraining and image generation tasks, respectively.}
    \label{fig:convergence_various_losses}
\end{figure*}

\begin{figure}[!t]
    \centering
    \includegraphics[width=.95\linewidth]{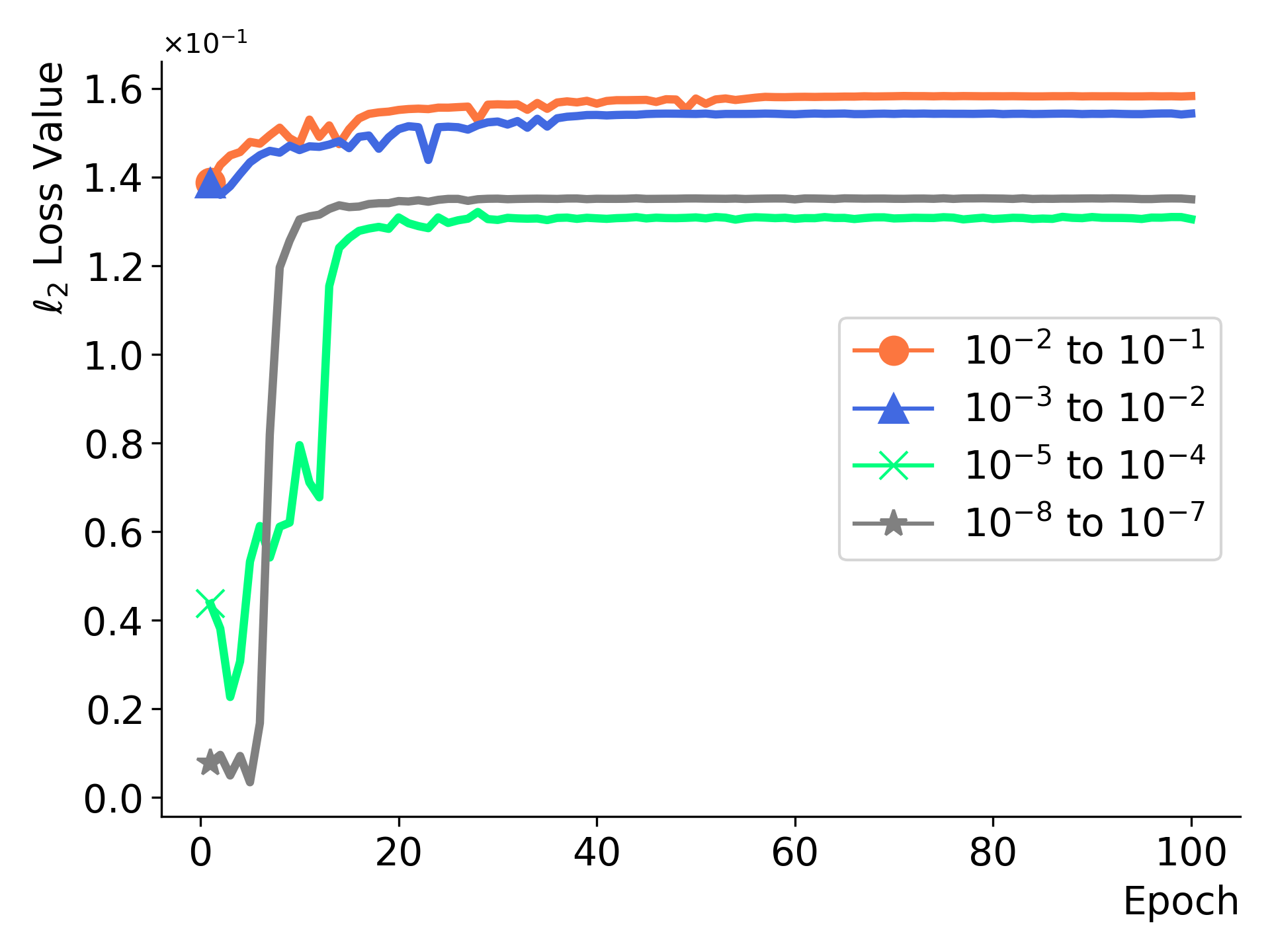}
    \vspace*{-6pt}
    \caption{Demonstration of the convergence behavior of the true loss $\|Z - W_0\|_2^2$ for different $P$ values using DGS-O on image generation tasks.}
    \label{fig:true_loss_dgso}
\end{figure}

\subsection{Convergence of Removal Loss}
We begin by evaluating the proposed DGSs in terms of their ability to hinder $\mathbb{R}$ from effectively learning to remove the embedded watermark, measured by the loss behavior during $\mathbb{R}$'s training process. With the implementation of DGSs, the attacker’s removal loss term $\mathbb{D}[\mathbb{R}(Y)]$ in (\ref{eq:r_loss}) is altered to $\mathbb{D}^\ast[\mathbb{R}(Y)]$, $\mathbb{D}[\widetilde{\mathbb{R}(Y)}]$, and $\overline{\mathbb{D}[\mathbb{R}(Y)]}$ for DGS-O, DGS-I, and DGS-L, respectively. Notably, all DGS variants are derived under the assumption of an $\ell_2$-norm removal loss. However, in practice, an attacker may adopt other forms of removal loss, such as the $\ell_1$-norm or a combination of $\ell_2$-norm and consistent loss, as introduced in \cite{zhang2020model}. This difference reflects a practical mismatch between what is assumed and what an actual attacker chooses, which we will also consider in our experiments. Throughout all settings, the fidelity loss is consistently defined using the $\ell_2$-norm, as shown in (\ref{eq:a_fidelity_loss}). The results are presented in Fig.~\ref{fig:convergence_various_losses} where the first and second rows correspond to DGS-O on deraining and image generation tasks, respectively, while the third and fourth rows are for DGS-I and DGS-L under the same conditions.

In both tasks, when no defense is applied, all loss values can be minimized to the range of $10^{-7}$ to $10^{-12}$, confirming the effectiveness of our gradient-based watermark removal attack. In contrast, once DGSs are implemented, none of the loss functions exhibit meaningful reductions. For DGS-O, the loss remains consistently around $10^{-2}$. The zoomed-in plots reveal a slight downward trend in the loss values; however, the changes are minimal, indicating that $\mathbb{R}$ fails to learn a reliable watermark removal under this defense. For DGS-I and all variants of DGS-L, the loss initially decreases but quickly rises and stabilizes near $10^{-2}$, preventing $\mathbb{R}$ from learning to remove the watermark. As shown in the zoom-in plots, the curves for DGS-I, DGS-L-S, DGS-L-M, and DGS-L-D are nearly identical and exhibit smooth convergence. In fact, DGS-I can be considered as a special case of DGS-L applied at the input layer. In addition, it can be seen from Fig. \ref{fig:convergence_various_losses} first and third columns that the proposed defenses, although designed with the $\ell_2$-norm removal loss, generalize effectively to cases involving alternative loss functions.

To further validate the ineffectiveness of training $\mathbb{R}$ under DGS-O, we visualize the loss between the raw output of $\mathbb{D}$, i.e., $Z$, and null watermark $W_0$ during the training process of $\mathbb{R}$ in the presence of DGS-O. The results are presented in Fig. \ref{fig:true_loss_dgso}, where the image generation task is used as an example. It can be seen that for all tested value ranges of $P$, the loss curves have increasing trends and eventually converge at a relatively high level above $1.2$, indicating that $\mathbb{R}$ is not able to remove the watermark. For DGS-I and all variants of DGS-L, the perturbations are constructed based on the proposed orthogonality mechanism. As a result, the discrepancy between the actual loss and the attacker’s removal loss becomes negligible and is hence not presented.

\begin{figure}[!t]
    \centering
    \includegraphics[width=1.0\linewidth]{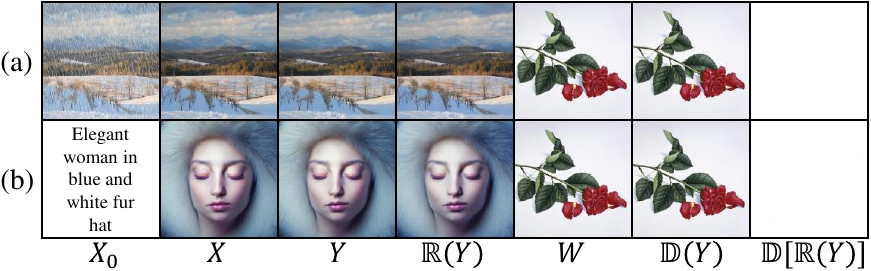}
    \caption{Visualization of the gradient-based watermark removal attack in the absence of defense. (a) Deraining. (b) Image Generation.}
    \label{fig:no_defense}
\end{figure}

\begin{figure}[!t]
    \centering
    \includegraphics[width=1.0\linewidth]{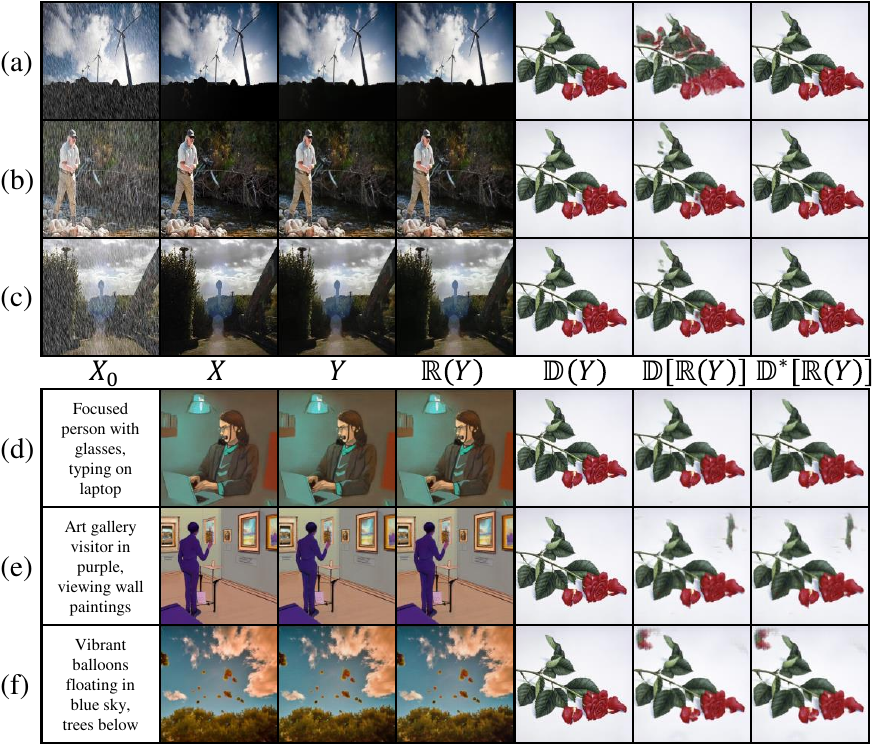}
    \caption{Demonstration of the gradient-based watermark removal attack under the proposed DGS-O defense. Subfigures (a)--(c) correspond to deraining, (d)--(f) refer to image generation. The watermark $W$ is consistent with that in Fig. \ref{fig:no_defense}. The attacker employs $\ell_1$ loss in (a) and (d), $\ell_2$ loss in (b) and (e), and a combination of $\ell_2$ and consistent loss in (c) and (f).}
    \label{fig:dgso_effectiveness}
\end{figure}

\subsection{Demonstrative Examples}
\begin{figure}[!t]
    \centering
    \includegraphics[width=1.0\linewidth]{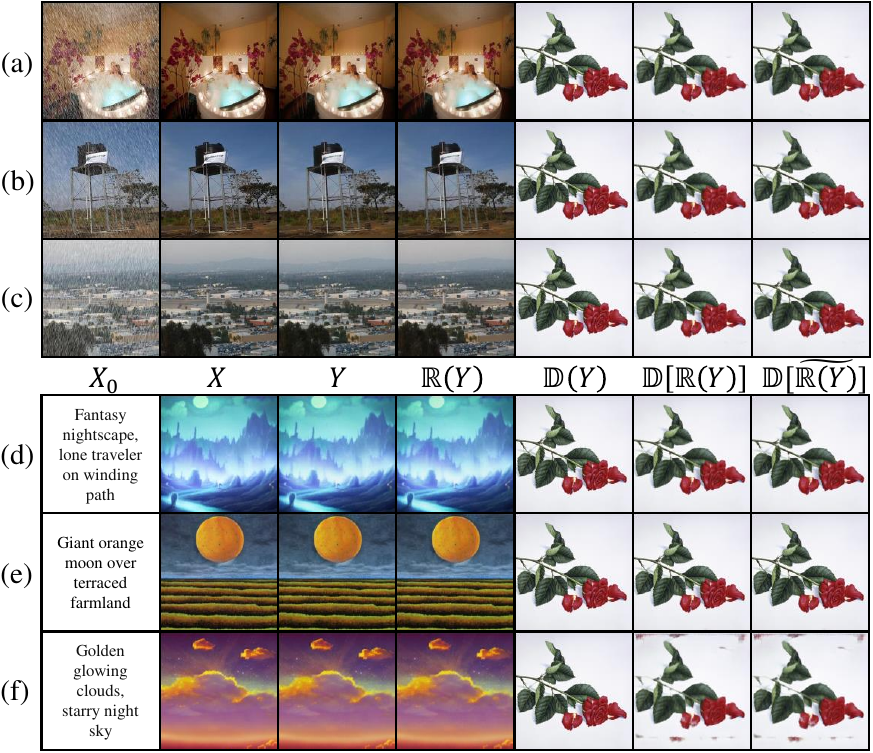}
    \caption{Demonstration of the gradient-based watermark removal attack under the proposed DGS-I defense. Subfigures follow the same layout as Fig. \ref{fig:dgso_effectiveness}.}  
    \label{fig:dgsi_effectiveness}
\end{figure}

\begin{figure}[!t]
    \centering
    \includegraphics[width=1.0\linewidth]{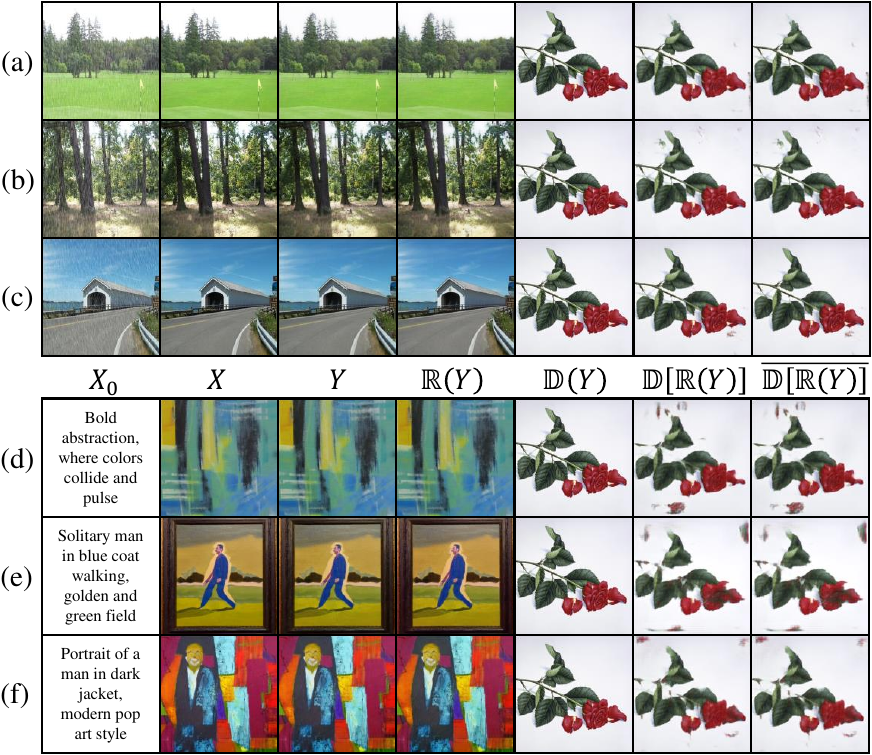}
    \caption{Demonstration of the gradient-based watermark removal attack under the proposed DGS-L-M defense. Subfigures follow the same layout as Fig. \ref{fig:dgso_effectiveness}.}
    \label{fig:dgsl_m_effectiveness}
\end{figure}

The demonstrative examples for both deraining and image generation tasks before and after the proposed DGS-O, DGS-I, and DGS-L-M are presented in Figs. \ref{fig:no_defense}, \ref{fig:dgso_effectiveness}, \ref{fig:dgsi_effectiveness} and \ref{fig:dgsl_m_effectiveness}, respectively. Images in each figure are drawn from: the original input $X_0$, the processed but non-watermarked $X$, the watermarked $Y$, the attacked $\mathbb{R}(Y)$, the ground-truth watermark $W$, the decoded watermark without attack $\mathbb{D}(Y)$, decoded watermark after attack $\mathbb{D}[\mathbb{R}(Y)]$ (equivalently denoted as $Z$), and the corresponding DGSs-perturbed decoding result ($\mathbb{D}^*[\mathbb{R}(Y)]$ for DGS-O, $\mathbb{D}[\widetilde{\mathbb{R}(Y)}]$ for DGS-I, and $\overline{\mathbb{D}[\mathbb{R}(Y)]}$ for DGS-L-M). It can be seen from Fig. \ref{fig:no_defense} last column that, in the absence of defense, the gradient-based remover effectively removes the watermark in both tasks, resulting in nearly blank decoded outputs from $\mathbb{D}$, while maintaining high image quality. The average PSNR values for the deraining and image generation tasks are $37.9523$ and $29.6088$, respectively, and the MS-SSIM values are $0.9968$ and $0.9806$. In contrast, the deployment of DGSs prevents this outcome. This resistance holds not only under matched loss function, as shown in subfigures (b) and (e) of Figs.\ref{fig:dgso_effectiveness}, \ref{fig:dgsi_effectiveness}, and \ref{fig:dgsl_m_effectiveness}, but also under mismatched loss functions, as evidenced by the remaining subfigures. Moreover, the comparisons of the last two columns show that, when DGSs are employed, the outputs are nearly indistinguishable from $\mathbb{D}[\mathbb{R}(Y)]$, maintaining high visual fidelity and enabling legitimate queries to function normally, while the learning process of $\mathbb{R}$ is effectively impeded.

\begin{figure}[!t]
    \centering
    \includegraphics[width=1.0\linewidth]{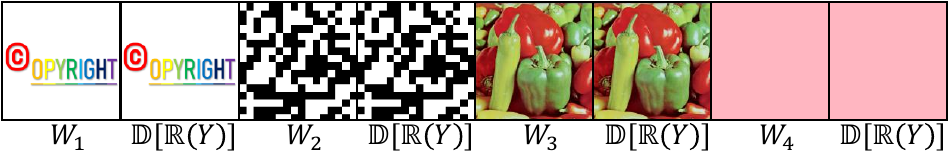}
    \caption{Demonstration of the tested marks for watermark sensitivity analysis, including text image ($W_1$), binary pattern ($W_2$), complex image ($W_3$), and solid color ($W_4$).} 
    \label{fig:watermark_sensitivity}
\end{figure}

\begin{figure}[!t]
    \centering
    \includegraphics[width=1.0\linewidth]{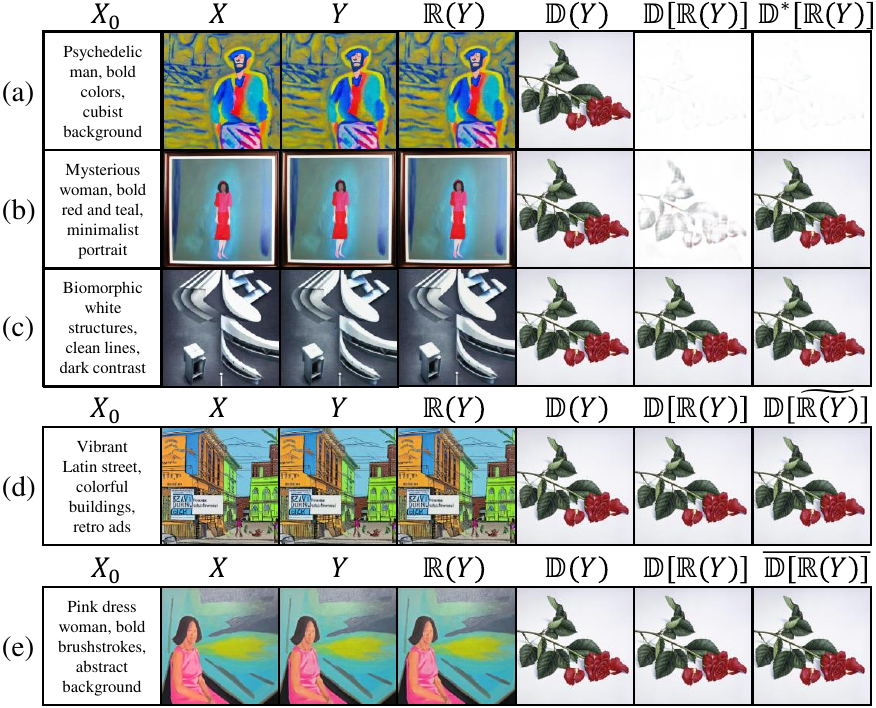}
    \caption{Demonstration of the robustness of the proposed DGSs under gradient sign flipping attacks. The same watermark $W$ as in Fig.~\ref{fig:no_defense} is used, with image generation task as an example. For DGS-O, $\Lambda_i$ is randomly sampled within (a) $[10^{-5}, 10^{-4}]$, (b) $[10^{-6}, 10^{-5}]$, (c) $[10^{-7}, 10^{-6}]$. Results for DGS-I and DGS-L are shown in (d) and (e), respectively.}
    \label{fig:attacker_flip}
\end{figure}

\begin{figure}[!t]
    \centering
    \includegraphics[width=1.0\linewidth]{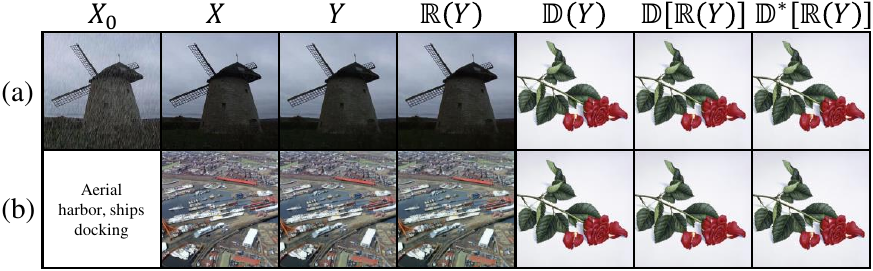}
    \caption{Demonstrative examples of DGS-O robustness against $Z$ restoration for (a) deraining and (b) image generation.}
    \label{fig:dgso_limitation}
\end{figure}
  
\subsection{Robustness}
\label{sec:robustness}
We first assess the sensitivity of the proposed DGSs across various $W$ including text image $W_1$, binary pattern $W_2$, complex image $W_3$, and solid color $W_4$. The demonstrative results are presented in Fig. \ref{fig:watermark_sensitivity}, which indicates high insensitivity. Furthermore, we examine the case where the attacker replaces $W_0$ in the removal loss (\ref{eq:r_loss}) by $W_1$ to $W_4$. While the attack succeeds in the absence of DGSs, it consistently fails when any DGS variant is applied.
  
In the case of gradient sign flipping, an attacker may attempt to reverse the sign of the removal loss gradient if the underlying defense is detected. While such reorientation may partially offset the protective effect, it cannot fully neutralize it. This partial compensation creates a limited opportunity for $\mathbb{R}$ to remove the watermark, though not most efficiently. Nevertheless, for DGS-O, the use of small $\Lambda_i$ values effectively suppresses the learning rate to hinder the attack. For DGS-I and DGS-L, the perturbations generated by the orthogonal mechanism orient the gradient with randomness to suppress the attack. The results are illustrated in Fig. \ref{fig:attacker_flip} with image generation task as an example. We note that, for DGS-O, when $\Lambda_i$ is within the interval $[10^{-6},10^{-5}]$ or smaller and for both DGS-I and DGS-L, DGSs remains robust to gradient-based removal attacks.

\begin{table}[!t]
  \centering
  \caption{Robustness test of DGS-O against JPEG compression, where $10^{-5}<\Lambda_i<10^{-4}$, PSNR is in dB, and $0\leq\text{MS-SSIM},\text{SR}\leq 1$.}
  \setlength{\tabcolsep}{5pt}
  \label{tab:JPEG_compression}
  \begin{tabular}{c|ccc|ccc}
  \hline
  \hline
  \multirow{2}{*}{Factor} & \multicolumn{3}{c|}{Deraining} & \multicolumn{3}{c}{Image Generation} \\
  \cline{2-7}
  & PSNR$\uparrow$ & MS-SSIM$\uparrow$ & SR$\uparrow$ & PSNR$\uparrow$ & MS-SSIM$\uparrow$ & SR$\uparrow$  \\
  \hline
  $10\%$ & $26.8063$ & $0.9559$ & $1.00$ & $26.8010$ & $0.9558$ & $1.00$ \\ 
  $20\%$ & $28.9928$ & $0.9738$ & $1.00$ & $28.9929$ & $0.9736$ & $1.00$ \\ 
  $30\%$ & $30.1931$ & $0.9828$ & $1.00$ & $30.1957$ & $0.9828$ & $1.00$ \\ 
  $40\%$ & $30.9914$ & $0.9862$ & $1.00$ & $30.9895$ & $0.9862$ & $1.00$ \\ 
  \hline
  \hline
  \end{tabular}
\end{table}

\begin{table}[!t]
  \centering
  \caption{Robustness test of DGS against WGN addition, where $10^{-5}<\Lambda_i<10^{-4}$, noise level and PSNR are in dB, and $0\leq\text{MS-SSIM},\text{SR}\leq 1$.}
  \setlength{\tabcolsep}{5pt}
  \label{tab:Gaussian_noise}
  \begin{tabular}{c|ccc|ccc}
  \hline
  \hline
  \multirow{2}{*}{\begin{tabular}{@{}c@{}} Noise \\ Level \end{tabular}} & \multicolumn{3}{c|}{Deraining} & \multicolumn{3}{c}{Image Generation} \\
  \cline{2-7}
  & PSNR$\uparrow$ & MS-SSIM$\uparrow$ & SR$\uparrow$ & PSNR$\uparrow$ & MS-SSIM$\uparrow$ & SR$\uparrow$  \\
  \hline
  $0$ & $1.7245$ & $0.3049$ & $1.00$ & $1.7270$ & $0.3053$ & $1.00$ \\ 
  $10$ & $11.7238$ & $0.5598$ & $1.00$ & $11.7238$ & $0.5597$ & $1.00$ \\ 
  $20$ & $21.7249$ & $0.8007$ & $1.00$ & $21.7256$ & $0.8005$ & $1.00$ \\ 
  $30$ & $31.7249$ & $0.9568$ & $1.00$ & $31.7268$ & $0.9567$ & $1.00$ \\ 
  \hline
  \hline
  \end{tabular}
  \end{table}

\begin{table}[!t]
  \centering
  \caption{Robustness test of DGS against lattice attack \cite{Liu2023Erase_Removal_Attack}, where $10^{-5}<\Lambda_i<10^{-4}$, PSNR is in dB, and $0< \text{MS-SSIM},\text{SR}<1$.}
  \setlength{\tabcolsep}{5pt}
  \label{tab:lattice_attack}
  \begin{tabular}{c|ccc|ccc}
  \hline
  \hline
  \multirow{2}{*}{Step} & \multicolumn{3}{c|}{Deraining} & \multicolumn{3}{c}{Image Generation} \\
  \cline{2-7}
  & PSNR$\uparrow$ & MS-SSIM$\uparrow$ & SR$\uparrow$ & PSNR$\uparrow$ & MS-SSIM$\uparrow$ & SR$\uparrow$  \\
  \hline
  $2$ & $12.2735$ & $0.6395$ & $1.00$ & $12.2579$ & $0.6398$ & $1.00$ \\ 
  $6$ & $21.7585$ & $0.8249$ & $1.00$ & $21.6827$ & $0.8234$ & $1.00$ \\ 
  $11$ & $26.8360$ & $0.9260$ & $1.00$ & $26.8659$ & $0.9274$ & $1.00$ \\ 
  $16$ & $30.3175$ & $0.9644$ & $1.00$ & $30.2486$ & $0.9635$ & $1.00$ \\ 
  \hline
  \hline
  \end{tabular}
  \end{table}
  
As discussed in Section \ref{sec:dgso_limitation_analysis}, DGS-O may be vulnerable against the attempt to restore $Z$ if the closed-form solution in (\ref{eq:ultimate_trans}) is known to the public. Meanwhile, its fixed applied location could be a loophole. To evaluate these, we consider $Z$ restoration and three types of post-processing that an attacker may apply:
\begin{itemize}
\item JPEG compression.
\item Noise addition.
\item Lattice attack \cite{Liu2023Erase_Removal_Attack}.
\end{itemize}

Among them, JPEG compression and additive white Gaussian noise are standard image degradation methods, whereas the lattice attack modifies pixels by assigning random values at fixed intervals for the purpose of watermark removal. The results of these attacks are shown in Figure~\ref{fig:dgso_limitation} and Tables~\ref{tab:JPEG_compression}--\ref{tab:lattice_attack}. For $Z$ restoration, it can be seen that the assumption of $P = I$ fails to compromise the defense. For post-processing strategies, even under severe degradation, such as a JPEG quality factor of $10\%$, $0$ dB noise level, and a lattice attack that randomly perturbs one out of every three pixels (step size of 2), the defense success rate remains at $100\%$. These results highlight the strong robustness of the proposed DGS-O. To further verify the robustness, let $F$ denote a generic additive perturbation introduced by the attacker’s post-processing, which is assumed to be independent of $Z$. Under this condition, Eq.~(\ref{eq:ultimate_trans}) is revised as:
\begin{equation}
    \label{eq:interfered_shield}
    Z^{\ast}_{\text{Interf}} = -PZ + (P+I)W + F.
\end{equation}
Accordingly, the gradient component used by an attacker, originally defined in Eq.~(\ref{eq:poison_gradient}), becomes:
\begin{align}
    & 2{\left( {Z^{\ast}_{\text{Interf}} - {W_0}} \right)^T}\frac{{\partial Z^\ast_{\text{Interf}}}}{{\partial \mathbb{R}(Y)}} \notag\\
    & = -2{\left[ {Z - \left({W_0} - F \right)} \right]^T}P\frac{{\partial Z}}{{\partial \mathbb{R}(Y)}}.
\end{align}

\noindent This result indicates that, with DGS-O in place, the additive perturbation $F$ shifts the optimization objective away from $W_0$ to $W_0 - F$. However, this does not fundamentally weaken the defense, as the core divergence mechanism remains effective. Moreover, the consistent performance of DGS-O across both deraining and image generation tasks suggests that its effectiveness is insensitive to the underlying data distribution, underscoring its generalizability to a broad range of generative models.

\begin{table}[!t]
  \centering
  \caption{Average Runtime for DGSs when the attacker employs the $\ell_2$ removal loss function, where DGS-O is measured in milliseconds while DGS-I and DGS-L are in seconds.}
  \setlength{\tabcolsep}{2pt} 
  \renewcommand{\arraystretch}{1.3} 
  \label{tab:runtime}
  \begin{tabular}{c|ccccc}
  \hline
  \hline
  Task & DGS-O & DGS-I & DGS-L-S & DGS-L-M & DGS-L-D\\ \hline
  Deraining & \textbf{$0.083$ ms} & $0.071$ s & $0.060$ s & $0.047$ s & $0.035$ s\\
  T2I Generation & \textbf{$0.133$ ms} & $0.067$ s & $0.058$ s & $0.046$ s & $0.033$ s\\
  \hline
  \hline
  \end{tabular}
\end{table}

\subsection{Computational Complexity}
\label{sec:compute_complex}
For $\mathbb{D}$ in CNN backbone with $l$ layers and $d \times d$ RGB image output, the computational complexities for DGS-O, DGS-I, and DGS-L (applied to the output of layer $k$) are $\mathcal{O}(d^2)$, $\mathcal{O}(ld^2)$, and $\mathcal{O}((l-k)d^2)$, respectively. We measure the average runtime of all methods under the same setting as the victim model \cite{Zhang2024Robust_Box_Free}, with results summarized in Table~\ref{tab:runtime}. All tests are conducted on a single NVIDIA RTX 5000 Ada Generation GPU (32GB GDDR6). Although all defenses introduce only a modest computational overhead, DGS-O stands out for its remarkably low latency, with runtimes of $0.083$ ms and $0.133$ ms for the two tasks, respectively, owing to its concise closed-form solution. Under a real service load, watermark verification is an independent, per-image operation, meaning the total computational cost scales linearly with verification requests and ensuring all proposed DGS variants impose an overhead well within an acceptable range for practical deployment.

\begin{table}[!t]
  \centering
  \caption{Guideline for choosing the appropriate DGS variant.}
  \label{tab:dgs_guideline}
  \setlength{\tabcolsep}{2pt} 
  \renewcommand{\arraystretch}{1.3} 
  \resizebox{\columnwidth}{!}{ 
  \begin{tabular}{c|cccc}
  \hline
  \hline
  Method & Location & Latency & Reliability & Application Scenario\\ \hline
  DGS-O & Output & Low & High & Universal (Scenario 1 \& 2)\\
  DGS-I & Input & High & Very High & Universal (Scenario 1 \& 2)\\
  DGS-L & Internal Layer & Moderate & Very High & Model Owner Only (Scenario 2)\\
  \hline
  \hline
  \end{tabular}
  }
\end{table}

\section{Further Discussion and Future Work}
To facilitate practical deployment, we provide a detailed guideline in Table~\ref{tab:dgs_guideline} for selecting the appropriate DGS variant based on factors such as defense location, latency, reliability, and application scenario, synthesized from the discussions in Section~\ref{sec:summary_dgss} and Section~\ref{sec:compute_complex}. Furthermore, it is important to clarify that our proposed DGS family is specifically tailored to mitigate gradient-based watermark removal attacks by disrupting the optimization of the remover. Consequently, attacks that completely bypass the decoder, such as those relying on training a separate surrogate model to approximate the watermark extraction process \cite{an2024box}, fall outside the scope of this defense. Developing countermeasures against such transferable removal attacks represents a critical and distinct avenue for future research.

\section{Conclusion}
Current box-free model watermarking for protecting generative models rely on a dedicated decoder $\mathbb{D}$ to extract watermarks directly from generated images. However, the inherent coupling between this decoder and the watermark encoder $\mathbb{E}$ presents a vulnerability: black-box access to $\mathbb{D}$ may expose the watermarking mechanism, which may be exploited for watermark removal. Building upon this insight, we have revealed a novel gradient-based watermark removal attack that effectively removes state-of-the-art box-free watermarks. To counteract this threat, we have proposed Decoder Gradient Shields (DGSs), a family of defenses implemented within the black-box API for watermark decoder, including DGS-O, DGS-I, and DGS-L. Among them, DGS-O has a closed-form solution that reorients and rescales gradients with a newly introduced positive definite matrix, guided by the proposed Gradient Norm Reduction and Randomness Injection mechanisms. For DGS-I and DGS-L, we have developed a novel scheme to generate perturbations orthogonal to the gradient component and apply them to the input or other layer outputs of watermark decoder. We have also provided provable performance against effective training of watermark remover, while preserving the image quality. Moreover, DGS-O, DGS-I, and DGS-L can be catered to different application scenarios, offering a set of practical defense solutions. Extensive experiments on both deraining and image generation tasks verify the robustness and practicality of our methods, providing a strong foundation for real-world deployment of secure and trustworthy DNN watermark decoders to protect the intellectual property of generative models.


\bibliographystyle{IEEEtran}
\bibliography{ref}

\begin{thebibliography}{10}
\providecommand{\url}[1]{#1}
\csname url@samestyle\endcsname
\providecommand{\newblock}{\relax}
\providecommand{\bibinfo}[2]{#2}
\providecommand{\BIBentrySTDinterwordspacing}{\spaceskip=0pt\relax}
\providecommand{\BIBentryALTinterwordstretchfactor}{4}
\providecommand{\BIBentryALTinterwordspacing}{\spaceskip=\fontdimen2\font plus
\BIBentryALTinterwordstretchfactor\fontdimen3\font minus \fontdimen4\font\relax}
\providecommand{\BIBforeignlanguage}[2]{{%
\expandafter\ifx\csname l@#1\endcsname\relax
\typeout{** WARNING: IEEEtran.bst: No hyphenation pattern has been}%
\typeout{** loaded for the language `#1'. Using the pattern for}%
\typeout{** the default language instead.}%
\else
\language=\csname l@#1\endcsname
\fi
#2}}
\providecommand{\BIBdecl}{\relax}
\BIBdecl

\bibitem{An2025Decoder}
H.~An, G.~Hua, Z.~Fang, G.~Xu, S.~Rahardja, and Y.~Fang, ``Decoder gradient shield: Provable and high-fidelity prevention of gradient-based box-free watermark removal,'' in \emph{Proceedings of the IEEE/CVF Conference on Computer Vision and Pattern Recognition (CVPR)}, 2025.

\bibitem{Gov_China_AI}
\BIBentryALTinterwordspacing
{Cyberspace Administration of China}, {National Development and Reform Commission}, {Ministry of Education}, {Ministry of Science and Technology}, {Ministry of Industry and Information Technology}, {Ministry of Public Security}, and {National Radio and Television Administration}, ``Interim measures for the management of generative artificial intelligence services,'' Cyberspace Administration of China Website, 7 2023. [Online]. Available: \url{https://www.gov.cn/zhengce/zhengceku/202307/content_6891752.htm}
\BIBentrySTDinterwordspacing

\bibitem{Gov_US_AI}
J.~R. Biden, ``Executive order on the safe, secure, and trustworthy development and use of artificial intelligence,'' 2023.

\bibitem{Gov_EU_AI}
{Council of the European Union}, ``Proposal for a regulation of the european parliament and of the council laying down harmonised rules on artificial intelligence (artificial intelligence act) and amending certain union legislative acts - analysis of the final compromise text with a view to agreement,'' 2024.

\bibitem{11127818_Yihang}
Y.~Tao, S.~Hu, Z.~Fang, and Y.~Fang, ``Directed-cp: Directed collaborative perception for connected and autonomous vehicles via proactive attention,'' in \emph{2025 IEEE International Conference on Robotics and Automation (ICRA)}, 2025, pp. 7004--7010.

\bibitem{hucpguard2025_Yihang}
S.~Hu, Y.~Tao, G.~Xu, Y.~Deng, X.~Chen, Y.~Fang, and S.~Kwong, ``Cp-guard: malicious agent detection and defense in collaborative bird's eye view perception,'' in \emph{Proceedings of the Thirty-Ninth AAAI Conference on Artificial Intelligence and Thirty-Seventh Conference on Innovative Applications of Artificial Intelligence and Fifteenth Symposium on Educational Advances in Artificial Intelligence}, ser. AAAI'25.\hskip 1em plus 0.5em minus 0.4em\relax AAAI Press, 2025.

\bibitem{tao2025gcpguardedcollaborativeperception_Yihang}
\BIBentryALTinterwordspacing
Y.~Tao, S.~Hu, Y.~Hu, H.~An, H.~Cao, and Y.~Fang, ``Gcp: Guarded collaborative perception with spatial-temporal aware malicious agent detection,'' 2025. [Online]. Available: \url{https://arxiv.org/abs/2501.02450}
\BIBentrySTDinterwordspacing

\bibitem{Stargate2025}
\BIBentryALTinterwordspacing
OpenAI, ``The stargate project,'' 2025. [Online]. Available: \url{https://openai.com/index/announcing-the-stargate-project/}
\BIBentrySTDinterwordspacing

\bibitem{google_watermarking_nature}
S.~Dathathri, A.~See, S.~Ghaisas, P.-S. Huang, R.~McAdam, J.~Welbl, V.~Bachani, A.~Kaskasoli, R.~Stanforth, T.~Matejovicova \emph{et~al.}, ``Scalable watermarking for identifying large language model outputs,'' \emph{Nature}, vol. 634, no. 8035, pp. 818--823, 2024.

\bibitem{pan2024markllm}
L.~Pan, A.~Liu, Z.~He, Z.~Gao, X.~Zhao, Y.~Lu, B.~Zhou, S.~Liu, X.~Hu, L.~Wen \emph{et~al.}, ``Markllm: An open-source toolkit for llm watermarking,'' \emph{arXiv preprint arXiv:2405.10051}, 2024.

\bibitem{Li2021A_Survey}
Y.~Li, H.~Wang, and M.~Barni, ``A survey of deep neural network watermarking techniques,'' \emph{Neurocomputing}, vol. 461, pp. 171--193, 2021.

\bibitem{Lv2023A_White_Box}
P.~Lv, P.~Li, S.~Zhang, K.~Chen, R.~Liang, H.~Ma, Y.~Zhao, and Y.~Li, ``A robustness-assured white-box watermark in neural networks,'' \emph{IEEE Transactions on Dependable and Secure Computing}, vol.~20, no.~6, pp. 5214--5229, 2023.

\bibitem{Cui2024Steganographic_White_Box}
Q.~Cui, R.~Meng, C.~Xu, and C.-H. Chang, ``Steganographic passport: An owner and user verifiable credential for deep model {IP} protection without retraining,'' in \emph{IEEE/CVF Conference on Computer Vision and Pattern Recognition (CVPR)}, Los Alamitos, CA, USA, 2024, pp. 12\,302--12\,311.

\bibitem{Tondi2024Robust_White_Box}
B.~Tondi, A.~Costanzo, and M.~Barni, ``Robust and large-payload dnn watermarking via fixed, distribution-optimized, weights,'' \emph{IEEE Transactions on Dependable and Secure Computing}, pp. 1--17, 2024.

\bibitem{Adi2018Turning_Black_Box}
Y.~Adi, C.~Baum, M.~Cisse, B.~Pinkas, and J.~Keshet, ``Turning your weakness into a strength: Watermarking deep neural networks by backdooring,'' in \emph{27th USENIX Security Symposium (USENIX Security 18)}, Baltimore, MD, 2018, pp. 1615--1631.

\bibitem{Huang2023What_Nonintrusive_GAN}
Z.~Huang, B.~Li, Y.~Cai, R.~Wang, S.~Guo, L.~Fang, J.~Chen, and L.~Wang, ``What can discriminator do? towards box-free ownership verification of generative adversarial networks,'' in \emph{2023 IEEE/CVF International Conference on Computer Vision (ICCV)}, 2023, pp. 4986--4996.

\bibitem{Zhang2022Deep_Box_Free}
J.~Zhang, D.~Chen, J.~Liao, W.~Zhang, H.~Feng, G.~Hua, and N.~Yu, ``Deep model intellectual property protection via deep watermarking,'' \emph{IEEE Transactions on Pattern Analysis \& Machine Intelligence}, vol.~44, no.~08, pp. 4005--4020, 2022.

\bibitem{Zhang2024Robust_Box_Free}
J.~Zhang, D.~Chen, J.~Liao, Z.~Ma, H.~Fang, W.~Zhang, H.~Feng, G.~Hua, and N.~Yu, ``Robust model watermarking for image processing networks via structure consistency,'' \emph{IEEE Transactions on Pattern Analysis and Machine Intelligence}, vol.~46, no.~10, pp. 6985--6992, 2024.

\bibitem{ilyas2018black_adversarial}
A.~Ilyas, L.~Engstrom, A.~Athalye, and J.~Lin, ``Black-box adversarial attacks with limited queries and information,'' in \emph{International conference on machine learning}, 2018, pp. 2137--2146.

\bibitem{Dong2022Query_adversarial}
Y.~Dong, S.~Cheng, T.~Pang, H.~Su, and J.~Zhu, ``Query-efficient black-box adversarial attacks guided by a transfer-based prior,'' \emph{IEEE Transactions on Pattern Analysis and Machine Intelligence}, vol.~44, no.~12, pp. 9536--9548, 2022.

\bibitem{shi2022query_adversarial}
Y.~Shi, Y.~Han, Q.~Hu, Y.~Yang, and Q.~Tian, ``Query-efficient black-box adversarial attack with customized iteration and sampling,'' \emph{IEEE Transactions on Pattern Analysis and Machine Intelligence}, vol.~45, no.~2, pp. 2226--2245, 2022.

\bibitem{yu2021artificial}
N.~Yu, V.~Skripniuk, S.~Abdelnabi, and M.~Fritz, ``Artificial fingerprinting for generative models: Rooting deepfake attribution in training data,'' in \emph{Proceedings of the IEEE/CVF International conference on computer vision}, 2021, pp. 14\,448--14\,457.

\bibitem{yu2022responsible}
N.~Yu, V.~Skripniuk, D.~Chen, L.~Davis, and M.~Fritz, ``Responsible disclosure of generative models using scalable fingerprinting,'' in \emph{Proc. ICLR}, 2022.

\bibitem{Wu2021Watermarking_Box_Free}
H.~Wu, G.~Liu, Y.~Yao, and X.~Zhang, ``Watermarking neural networks with watermarked images,'' \emph{IEEE Transactions on Circuits and Systems for Video Technology}, vol.~31, no.~7, pp. 2591--2601, 2021.

\bibitem{Lukas2023PTW_Box_Free}
N.~Lukas and F.~Kerschbaum, ``{PTW}: Pivotal tuning watermarking for pre-trained image generators,'' in \emph{32nd USENIX Security Symposium (USENIX Security 23)}, 2023, pp. 2241--2258.

\bibitem{Lin2024A_Box_Free}
D.~Lin, B.~Tondi, B.~Li, and M.~Barni, ``A {CycleGAN} watermarking method for ownership verification,'' \emph{IEEE Transactions on Dependable and Secure Computing}, pp. 1--15, 2024.

\bibitem{Fei2024Wide_Box_Free}
J.~Fei, Z.~Xia, B.~Tondi, and M.~Barni, ``Wide flat minimum watermarking for robust ownership verification of {GANs},'' \emph{IEEE Transactions on Information Forensics and Security}, vol.~19, pp. 8322--8337, 2024.

\bibitem{Barbalau2020Black}
A.~Barbalau, A.~Cosma, R.~T. Ionescu, and M.~Popescu, ``Black-box ripper: Copying black-box models using generative evolutionary algorithms,'' in \emph{Advances in Neural Information Processing Systems}, vol.~33, 2020, pp. 20\,120--20\,129.

\bibitem{Kariyappa2021Maze}
S.~Kariyappa, A.~Prakash, and M.~K. Qureshi, ``Maze: Data-free model stealing attack using zeroth-order gradient estimation,'' in \emph{2021 IEEE/CVF Conference on Computer Vision and Pattern Recognition (CVPR)}, 2021, pp. 13\,809--13\,818.

\bibitem{Ma2023DivTheft}
Z.~Ma, X.~Liu, Y.~Liu, X.~Liu, Z.~Qin, and K.~Ren, ``Divtheft: An ensemble model stealing attack by divide-and-conquer,'' \emph{IEEE Transactions on Dependable and Secure Computing}, vol.~20, no.~6, pp. 4810--4822, 2023.

\bibitem{Sha2023Cant}
Z.~Sha, X.~He, N.~Yu, M.~Backes, and Y.~Zhang, ``Can't steal? cont-steal! contrastive stealing attacks against image encoders,'' in \emph{2023 IEEE/CVF Conference on Computer Vision and Pattern Recognition (CVPR)}, Los Alamitos, CA, USA, Jun. 2023, pp. 16\,373--16\,383.

\bibitem{Liu2023Erase_Removal_Attack}
H.~Liu, T.~Xiang, S.~Guo, H.~Li, T.~Zhang, and X.~Liao, ``Erase and repair: An efficient box-free removal attack on high-capacity deep hiding,'' \emph{IEEE Transactions on Information Forensics and Security}, vol.~18, pp. 5229--5242, 2023.

\bibitem{goodfellow2014explaining}
I.~J. Goodfellow, J.~Shlens, and C.~Szegedy, ``Explaining and harnessing adversarial examples,'' \emph{arXiv preprint arXiv:1412.6572}, 2014.

\bibitem{madry2018towards}
A.~Madry, A.~Makelov, L.~Schmidt, D.~Tsipras, and A.~Vladu, ``Towards deep learning models resistant to adversarial attacks,'' in \emph{International Conference on Learning Representations}, 2018.

\bibitem{Mantas2022How}
M.~Mazeika, B.~Li, and D.~A. Forsyth, ``How to steer your adversary: Targeted and efficient model stealing defenses with gradient redirection,'' in \emph{International Conference on Machine Learning, {ICML} 2022, 17-23 July 2022, Baltimore, Maryland, {USA}}, vol. 162, 2022, pp. 15\,241--15\,254.

\bibitem{pp_2020_iclr_defense_model_stealing}
T.~Orekondy, B.~Schiele, and M.~Fritz, ``Prediction poisoning: Towards defenses against dnn model stealing attacks,'' \emph{arXiv preprint arXiv:1906.10908}, 2019.

\bibitem{Sanyal2022Towards_Steal_Attack}
S.~Sanyal, S.~Addepalli, and R.~V. Babu, ``Towards data-free model stealing in a hard label setting,'' in \emph{2022 IEEE/CVF Conference on Computer Vision and Pattern Recognition (CVPR)}, 2022, pp. 15\,263--15\,272.

\bibitem{Haykin2002Adaptive}
S.~Haykin, \emph{Adaptive filter theory}, 4th~ed.\hskip 1em plus 0.5em minus 0.4em\relax Prentice Hall, 2002.

\bibitem{Zhang2023Categorical}
H.~Zhang, G.~Hua, X.~Wang, H.~Jiang, and W.~Yang, ``Categorical inference poisoning: Verifiable defense against black-box dnn model stealing without constraining surrogate data and query times,'' \emph{IEEE Transactions on Information Forensics and Security}, vol.~18, pp. 1473--1486, 2023.

\bibitem{everingham2010pascal_Dataset_Pascal}
M.~Everingham, L.~Van~Gool, C.~K. Williams, J.~Winn, and A.~Zisserman, ``The pascal visual object classes (voc) challenge,'' \emph{International journal of computer vision}, vol.~88, pp. 303--338, 2010.

\bibitem{zhang2018density}
H.~Zhang and V.~M. Patel, ``Density-aware single image de-raining using a multi-stream dense network,'' in \emph{Proceedings of the IEEE conference on computer vision and pattern recognition}, 2018, pp. 695--704.

\bibitem{rombach2022high_Victim_StableDiffusion}
R.~Rombach, A.~Blattmann, D.~Lorenz, P.~Esser, and B.~Ommer, ``High-resolution image synthesis with latent diffusion models,'' in \emph{Proceedings of the IEEE/CVF conference on computer vision and pattern recognition}, 2022, pp. 10\,684--10\,695.

\bibitem{wang2003multiscale}
Z.~Wang, E.~P. Simoncelli, and A.~C. Bovik, ``Multiscale structural similarity for image quality assessment,'' in \emph{The Thrity-Seventh Asilomar Conference on Signals, Systems {\&} Computers}, vol.~2, 2003, pp. 1398--1402.

\bibitem{ronneberger2015u_UNet}
O.~Ronneberger, P.~Fischer, and T.~Brox, ``U-net: Convolutional networks for biomedical image segmentation,'' in \emph{Medical Image Computing and Computer-Assisted Intervention--MICCAI 2015: 18th International Conference, Munich, Germany, October 5-9, 2015, Proceedings, Part III 18}.\hskip 1em plus 0.5em minus 0.4em\relax Springer, 2015, pp. 234--241.

\bibitem{zhang2020model}
J.~Zhang, D.~Chen, J.~Liao, H.~Fang, W.~Zhang, W.~Zhou, H.~Cui, and N.~Yu, ``Model watermarking for image processing networks,'' in \emph{Proceedings of the AAAI conference on artificial intelligence}, vol.~34, no.~07, 2020, pp. 12\,805--12\,812.

\bibitem{an2024box}
H.~An, G.~Hua, Z.~Lin, and Y.~Fang, ``Box-free model watermarks are prone to black-box removal attacks,'' \emph{arXiv preprint arXiv:2405.09863}, 2024.

\end{thebibliography}

\end{document}